\documentclass[review]{elsarticle}

\usepackage{graphicx}
\usepackage{amsmath}
\usepackage{amssymb}
\usepackage{tabulary}
\usepackage{multirow}
\usepackage{float}
\usepackage{soul}
\usepackage{algorithm,algpseudocode}
\usepackage{hyperref}

\journal{Journal of \LaTeX\ Templates}

\begin{document}

\begin{frontmatter}

\title{Object Topological Character Acquisition by Inductive Learning}
%\tnotetext[mytitlenote]{Fully documented templates are available in the elsarticle package on \href{http://www.ctan.org/tex-archive/macros/latex/contrib/elsarticle}{CTAN}.}

%% Group authors per affiliation:
%\author{Wei Hui\fnref{he}}
%\author{Liping Yu\fnref{me}}
%\author{Yiran Wei\fnref{her}}
\author[lab]{Wei Hui}
\author[lab]{Liping Yu}
\author[lab1]{Yiran Wei}
\address[lab]{Laboratory of Algorithm for Cognitive Model, School of Computer Science, \\
Shanghai Key Laboratory of Data Science, Fudan University}
\address[lab1]{Shanghai Pinghe School}

%\author{Elsevier\fnref{myfootnote}}
%\address{Radarweg 29, Amsterdam}
%\fntext[myfootnote]{Since 1880.}
%
%%% or include affiliations in footnotes:
%\author[mymainaddress,mysecondaryaddress]{Elsevier Inc}
%\ead[url]{www.elsevier.com}
%
%\author[mysecondaryaddress]{Global Customer Service\corref{mycorrespondingauthor}}
%\cortext[mycorrespondingauthor]{Corresponding author}
%\ead{support@elsevier.com}
%
%\address[mymainaddress]{1600 John F Kennedy Boulevard, Philadelphia}
%\address[mysecondaryaddress]{360 spinePark Avenue South, New York}

\begin{abstract}
Understanding the shape and structure of objects is undoubtedly extremely important for object recognition, but the most common pattern recognition method currently used is machine learning, which often requires a large number of training data. The problem is that this kind of object-oriented learning lacks a priori knowledge. The amount of training data and the complexity of computations are very large, and it is hard to extract explicit knowledge after learning. This is typically called `knowing how without knowing why'. We adopted a method of inductive learning, hoping to derive conceptual knowledge of the shape of an object and its formal representation based on a small number of positive examples. It is clear that implementing object recognition is not based on simple physical features such as colors, edges, textures, etc., but on their common geometry, such as topologies, which are stable, persistent, and essential to recognition. In this paper, a formal representation of topological structure based on object’s skeleton (RTS) was proposed and the induction process of 'seeking common ground' is realized. This research helps promote the method of object recognition from empiricism to rationalism
%Understanding the shape and structure of objects is undoubtedly extremely important for object recognition, but the most common pattern recognition method currently used is machine learning, which often requires a large number of training data. However, this process actually involves over-learning by memorizing large amounts of data. The problem is that this kind of object-oriented learning lacks a priori knowledge. The amount of training data and the amount of calculations are very large, and it is impossible to extract explicit knowledge after learning. This is typically called 'knowing how without knowing why'. We adopted a method of inductive learning, hoping to derive conceptual knowledge of the shape of an object and its formal representation based on a small number of positive examples. It is clear that implementing object categorization is not based on simple physical features such as colors, edges, textures, etc., but on their common geometry, such as topologies, which are stable, persistent, and essential to categorization. In this paper, skeleton structure information based on the shape of the object was proposed to form the appropriate RTS (a formal representation of topological structure based on object's skeleton) characterization and the induction process of 'maximizing common points while minimizing differences,' and the generalization of the common features of the same class is realized. This research helps promote the method of object recognition from empiricism to rationalism.
\end{abstract}

\begin{keyword}
%\texttt{elsarticle.cls}\sep \LaTeX\sep Elsevier \sep template
Topological Character Acquisition\sep Inductive Learning\sep A Representation of Topological Structure Based on Object's Skeleton
%\sep Skeleton 
%\sep Skeleton Tree 
%Knowledge of Object Structure
%\MSC[2010] 00-01\sep 99-00
\end{keyword}

\end{frontmatter}

%\linenumbers

\section{Introduction}

The ultimate goal of computer vision is to give machines the ability to perceive and understand human vision. Object recognition and classification is a classic problem in computer vision. After deep-learning method has been introduced, object recognition seems no longer to be a challenging problem. As long as there are enough training data, and a lot of training and fine-tuning processes, it seems that object recognition can always achieve very high accuracy on a specific dataset.

Accordingly, does this mean that the object recognition problem has been solved? We do not think so because machine learning simplifies the identification problem into a classification problem for the bounding box. This does not reflect the actual processing of the object we recognize. The bounding box is often a circumscribed rectangular area. Although algorithm gives the probability that the rectangle contains positive cases, this is ambiguous because it does not know which pixels in the rectangle are the target objects and why they are positive examples. Although machine learning methods can make the precision rate very high, the costs of sample collection, labeling, and training are also very high. Such defects do not exist in shape-based object recognition methods, but these methods face a series of problems such as dimensional changes, viewing angle changes, shape deformation, occlusion, and defects; moreover, shape extraction is strongly influenced by the physical characteristics of the image. Although there are a lot of problems with the shape-based recognition strategy, we think it is closer to realitymore reasonable than the appearance-based machine learning recognition strategy. Cognitive scientific evidence in this area is very powerful.

%\paragraph{Installation} If the document class \emph{elsarticle} is not available on your computer, you can download and install the system package \emph{texlive-publishers} (Linux) or install the \LaTeX\ package \emph{elsarticle} using the package manager of your \TeX\ installation, which is typically \TeX\ Live or Mik\TeX.
%
%\paragraph{Usage} Once the package is properly installed, you can use the document class \emph{elsarticle} to create a manuscript. Please make sure that your manuscript follows the guidelines in the Guide for Authors of the relevant journal. It is not necessary to typeset your manuscript in exactly the same way as an article, unless you are submitting to a camera-ready copy (CRC) journal.
%
%\paragraph{Functionality} The Elsevier article class is based on the standard article class and supports almost all of the functionality of that class. In addition, it features commands and options to format the
%
%\item document style
%\item baselineskip
%\item front matter
%\item keywords and MSC codes
%\item theorems, definitions and proofs
%\item lables of enumerations
%\item citation style and labeling.
%\end{itemize}

\subsection{Evidence From Cognitive Psychology}
Cognitive psychologists have done extensive research and experiments on object recognition for infants under one year old \cite{Dueker2005Infants,Bloom2004Descartes,gopnik1999scientist,Shinskey2014Picturing}. One of these experiments conducted by Dueker et al. \cite{Dueker2005Infants} displayed two dissimilar objects placed contiguous to one another to infants between four and six months old. Without prior experiences, infant participants did not indicate whether they identify what they see as one item as a whole or two items placed together. However, when researchers showed the infants one object from the display before the test and repeated the same experiment, the infants demonstrated that they are able to distinguish the boundary between the two objects. In other words, since the infants were already shown one of the objects and recognized that exact object during the later test, they discerned what they were seeing as two items touching one another instead of as a single item only. When objects similar to one of the two objects in the display were shown previously to the infants (for example, different types of boxes), the results were as follows. When the infants saw one similar object in advance, they were still unable to distinguish the boundary between the objects used in the test. However, when the infants were familiarized with up to three similar objects beforehand, they could well recognize the similar object as an individual item by itself, subsequently understanding what was in front of them as two items placed together instead of one item as a whole. Even if the familiarization process was done 72 hours before the actual test, infants between four and six months could still correctly segregate the objects in the display with three examples \cite{Dueker2005Infants}.

This experiment provided strong evidence that even extremely young infants possess a generalization ability that requires only a small sample size. This conclusion was also reached in a range of similar experiments. For instance, Bloom et al. \cite{Bloom2004Descartes} showed that five-month-old infants are able to relate life-sized pictures to their corresponding real-world objects, facilitating a quick recognition for both familiar and novel objects. Gopnik et al. \cite{gopnik1999scientist} reported that infants are able to learn the use and nature of an unfamiliar object by merely a limited number of observations of adults, and can retain a precise recognition of this object even after a week. Shinskey et al. \cite{Shinskey2014Picturing} performed experiments confirming that infants nine months old could not only recognize items by learning their pictures but also understand the concept of object constancy for known objects.

All of the experiments mentioned above show infants’ extraordinary ability to learn the appearance of objects and their flexibility in recognizing them. 

\subsection{Evidence From Neurobiology}
The neurobiology community has conducted extensive experimental studies on neurons in the inferotemporal (IT) cortex of the brain and found that these neurons are capable of selectively responding to complex objects. These neurons are capable of expressing relatively high-level information about certain types of objects, and they have a response invariance to multiple changes in the object \cite{brincat2004underlying,Ito1995Size,Majaj2015Simple}. The most representative example is the discovery of so-called 'face cells' in the FFA region, which can specifically respond to real faces, monkey faces, or graphic patterns similar to face topology \cite{Bruce1981Visual,tsao2003faces,Tsao2006A}. On the issue of concept extraction, which is more advanced than direct visual stimulation, Lin et al. \cite{Lin2007Neural} studied neurons in the mouse hippocampus. Large-scale recordings suggested that the hippocampus is an intrinsic part of the hierarchical structure for generating concepts and knowledge in the brain. In contrast to the shape constancy of face cell responses, nest cell responses are independent of geometric shape, design style, color, odor, or construction materials. These systematic organizations revealed that the neural encoding of the abstract concept of nests is tuned toward the functional features of nests. These cells have a response invariance to changes in a nest’s physical shape, style, color, odor, or construction materials, and are not affected by locations and environments. This indicates that these cells respond selectively to the the abstract semantics of nest. In the study of cognitive neurobiological mechanisms on how the brain characterizes concepts, scholars have found that specific cortical regions and their connections through white matter have the ability to characterize higher-order semantic information \cite{Fang2018Semantic,Fernandino2015Concept,martin2016grapes}. Results of fMRI studies on how the brain represents object attributes and object categories and results of neuroanatomical and neurophysiological studies on monkeys both suggest that there is a large-scale intrinsic circuit loop and dynamic connections in the brain and they encode an object’s distinctive attributes and attribute combinations. This seems to imply that a flat graph structure is more likely to be a technical feasible way for implementation of object representation \cite{martin2016grapes}.

None of the above neurobiological findings indicates that the brain has an intrinsic mechanism for high-order feature extraction and complex pattern characterization of input stimuli. The basis for a selective response of a particular region of neurons is not the physical appearance of an object, but higher-level semantic information. Furthermore, the ability of category-generalization does not require an extensive sample training, and it seems that the geometric, topological, and even functional features of the concept can be grasped based on a small number of examples. It seems that biological systems naturally have the ability to 'extrapolate' and 'look beyond the appearance'. This is both a need for survival and a principle of evolutionary economics.

\subsection{Current Problems Encountered in the Engineering Field}
The above experimental evidence from neurobiology and cognitive psychology reflects the ability and efficiency of human object recognition. At present, there is a significant performance gap between the most mainstream machine learning-based object recognition methods. The current mainstream approach has the following characteristics.

First, it is data driven and relies on massive amounts of data. Most of the data is unlabeled, and it is not realistic to mark it, which requires a lot of manpower.

Second, the interpretability is poor. The classifier formed by machine learning is often a black box, and it is difficult to explain why a sample belongs to a positive case, or does not belong, because the classification weakens the analysis of the geometric and topological features of the object.

Third, it is a way of over-learning. In order to make the training sample cover all possible positive examples and combinations with various backgrounds as much as possible, a large number of positive and negative examples are produced for overtraining.

Fourth, the model has high complexity, difficult optimization, and poor generalization ability. For the current mainstream deep learning method, there is no upper limit on the neural network’s capacity and strong expression ability, which is an important advantage. However, this also causes great difficulty in the optimization of the model. The more complex the model, the more difficult it is to optimize the model. At the same time, the optimization of neural networks currently involves a large number of adjustments. Most of these adjustments are empirical and lack sufficient theoretical basis. Meanwhile, the complexity of the model also reduces its generalization ability to a certain extent. Usually, a model is only adapted to a specific dataset, and it is often necessary to reconstruct the model for data of other scenarios.

%Fifth, the calculation strength is high, the hardware requirements are high, and it is necessary to use hardware such as a GPU to train. Such high-cost results are often a large number of parameter values ​​with high complexity, and there is no explicit or concise description of the essential features of the class.

Fifth, the calculation strength is high, the hardware requirements are high, and it is necessary to use hardware such as a GPU to train. Such high-cost results are often a large number of parameter values with high complexity, and there is no explicit or concise description of the essential features of the class.

We are not satisfied that the recognition algorithm only outputs 'what'; we also want the recognition algorithm to explain 'why.' That is, we want to know what the model learned after learning a bunch of samples, not just a mapping of pictures to numbers (categories) through a series of complex nonlinear mappings. We need a way to generate an explicit connotation representation of a category that explains in detail why the new sample belongs to or does not belong to the category. We hope that the model can summarize the general patterns of similar objects after learning some examples. For example, mammals usually have four legs. We believe that higher levels of geometric and topological features are the key foundations for object recognition and classification. No matter what the weather, light, or distance, we can still easily identify cars of different shapes and sizes on the street. Because the structural knowledge of the car is the most important, we tend to use four wheels to describe the car rather than other low-level features, such as color, texture. We need to describe an object with a more essential, stable, and lasting feature.

\subsection{Organization of This Paper}
In this paper, we proposed a formal representation of topological structure based on object's skeleton (RTS) and introduced a spine-like axis (SPA) to further constrain the spatial distribution of the object structure. Experimental results showed that this method can effectively characterize the structural features of objects. Based on RTS, we replaced the Euclidean distance with the Fr{\'e}chet distance \cite{frechet1906quelques} to improve the previous skeleton path similarity measurement method. The OSB algorithm was used to elastically match the object structure to obtain the structural matching relationship between objects. Finally, we proposed a topological character acquisition method based on inductive learning that could effectively extract the display connotation representation of the category, which explains in detail why the new sample belongs to or does not belong to the category.

The remainder of the paper is organized as follows: Section 2 reviews the related work. Next, Section 3 describes the formal representation of topological structure based on object's skeleton (RTS). Then, Section 4 extends the efficient approach of using the similarity measure between two objects. Section 5 describes the topological character acquisition based on inductive learning. The experimental results of the proposed technique are shown in Section 6. Finally, a conclusion is given in Section 7.

\section{Related Work}
Forming a formal description of the geometry and topological features of the same type of object is the precondition of object recognition tasks, so that background changes, perspective changes, illumination changes, dimensional changes, and joint rotation can be dealt with. One of the representative methods following this shape characterization strategy is template matching. Generally, the template of the target to be identified is matched with all the samples in the training sample set, and a large number of samples containing many changes need to be stored \cite{Belongie2002Shape, Latecki2000Shape}. However, there are geometric deformations such as rotation and scaling between the actual target and the standard template. At the same time, due to structural changes, coupling methods, occlusion, illumination, etc., severe nonlinear variability causes many existing object recognition methods based on template matching to be powerless \cite{Sun2005Classification}. Bag-of-Words (BoW) \cite{sivic2003video,csurka2004visual} is proposed for the problem that the overall matching can not deal with occlusion, nonlinear distortion, and large intra-class changes of objects. It uses a learned set of higher frequency primitives as a dictionary, and uses a combined vector of primitive labels to represent the object. One of the main drawbacks of the BoW method is that the primitives are often independent of each other, losing relative spatial position information, and lacking the modeling of the constraint relationship between primitives; therefore, the order relationship of each primitive in space cannot be characterized. The spring model proposed by Pictorial Structure (PS) \cite{felzenszwalb2000efficient,felzenszwalb2003pictorial}, the relationship between the object parts is represented by a telescopic spring, and the model representation of the relationship between the features, as well as the star structure, the hierarchical structure, the tree structure, and the like. This was an effective representation of the geometric or topological features of an object. The method of syntactic analysis based on object structure is very suitable for some demand of interpretation, and can decompose many complex patterns into simple sub-patterns that are then divided into several primitives. Each target can be composed of primitives based on a certain relationship. By identifying the primitives, the sub-pattern is further identified, and the complex pattern is finally identified. This method is suitable for structurally strong modes, but extracting primitives is difficult. In order to obtain the grammar that is compatible with the pattern class, it is necessary to collect enough training pattern samples in advance, and then extract the corresponding grammar through the primitive extraction. This method still presents some difficulty for practical applications.

The emergence of deep learning methods brought the correct rate of object recognition to an unprecedented height, but its shortcomings are also obvious: they lack effective learning theory, require a large number of labeled training samples, have too many parameters, easily produce over-learning, and lack interpretability. The entire learning process lacks an induction of high-order features. It is worth emphasizing that deep learning is used for object recognition in order to improve accuracy, not to improve descriptiveness and interpretability.

The ultimate goal of computer vision is to enable machines to have visual perception and understanding like humans. Many attributes of objects can be used to identify and classify, for example, shape, color, texture, and brightness. The shape is usually represented by a binary image of the target range. Unlike other low-level features such as textures and colors, the shape is the most essential feature of the target object and has the basic structure represented by the target There are generally two types of methods for the characterization of shapes: one is a contour-based approach and the other is a region-based approach. Generally, the contour-based shape feature description method only considers the contour information of the shape, and extracts the contour feature information to describe the shape more in-line with human physiological vision. The Fourier descriptor (FD) \cite{zahn1972fourier} and the wavelet descriptor \cite{chuang1996wavelet} are two classic contour-based shape characterization methods.. However, the former does not provide local information of the shape and is sensitive to noise. The latter is very dependent on the starting point of the shape contour curve. The wavelet coefficients are susceptible to noise interference, and there is no translation, rotation, or scaling without deformation. Region-based shape descriptors examine all the pixels in the closed contour, can better express the features represented by the shape, and have obvious advantages for resisting contour noise and the like. Typical region-based structuring methods include the convex hull \cite{davies2004machine,batchelor1980hierarchical} and skeleton \cite{blum1973biological,calabi1968shape}. The skeleton is an effective structural descriptor. It can reflect the structure and topological features of the object, and form a simple shape representation. This description is highly flexible, fault-tolerant, and interpretable.

We propose an object structure representation method based on skeleton tree. For an object with skeleton information extracted, each branch of the skeleton is used as a primitive of the object structure, thereby decomposing the complex pattern of the original image into several simple sub-patterns. At the same time, the skeleton structure ensures a certain topological relationship between each primitive.We used the skeleton branches as the primitives, and extracted the structural features of the object hierarchically, which facilitated learning the essential information inherent in more objects from only a few samples. Other researchers have proposed some skeleton-based methods to learn the skeleton abstraction pattern of the class from similar samples. Demirci et al. \cite{Fatih2009Skeletal} built a similar skeleton prototype based on the many-to-many mapping relationship between skeleton nodes. However, in each step of the pairwise abstraction, the average strategy was chosen. This method had a preference for the new model to be added later, and was not good for evaluating the structural changes of similar objects. Torsello et al. \cite{Torsello2006Learning} found a joint attribute tree model that can best explain the same kind of samples through the minimum coding criterion, but this optimization process tends to converge to the local optimum. Shen et al. \cite{Wei2013Shape} proposed a multi-level clustering algorithm based on a joint common skeleton map; it could capture intra-class changes well, but its purpose was to improve the clustering algorithm. According to the common structure skeleton diagram obtained by clustering, the similarity between more robust shape clusters was obtained, and the final result was a graph structure, which did not make semantics emerging. Due to the complex scenes in the natural environment, we ignored low-level features such as color and texture, and only explored the shape representation of objects in binary images. We tried to give a semantic induction about the high-level features of similar objects so as to carry out some interpretability work. This research helps promote the method of object recognition from empiricism to rationalism.

\section{A Formal Representation of the Topological Structure Based on the Object's Skeleton (RTS).}
\label{sec:shape_res}

%We proposed a formal representation of the topological structure based on the object's skeleton (RTS). The main idea of ​​this method was to emphasize the structural properties of an object based on the theory of cognitive psychology. First, the object was decomposed into several connected components based on the skeleton, and each component was represented by a skeleton branch. The statistical characteristics of the skeleton branches were then taken: area ratio, length ratio, and skeleton path mass distribution.

We proposed a formal representation of the topological structure based on the object's skeleton (RTS). The main idea of this method was to emphasize the structural properties of an object based on the theory of cognitive psychology. First, the object was decomposed into several connected components based on the skeleton, and each component was represented by a skeleton branch. The statistical characteristics of the skeleton branches were then taken: area ratio, length ratio, and skeleton path mass distribution.

The skeleton is an effective descriptor of the structure of the object. It not only contains contour information, but also the topological structure of the target shape. It has the characteristics of sparse data and accurate shape. More importantly, the inscribed circle radius value and the skeleton path length of the skeleton feature point do not change with movement of the target’s limbs. Therefore, skeleton-based shape representation is very suitable for dealing with the problem of flexible changes and partial occlusion of the target. By using vectors to record all the information on the skeleton, we can retain the main part of the skeleton information and also the details. Converting a graph into a vector is important in image processing. In this paper, we examined the stability of the root node, and transformed the skeleton map into a skeleton tree with a height of one. Each skeleton branch from the root node to the end node was a visual component of the object. We further calculate the feature quantity to transform the skeleton branches into vectors, and constrain the relative spatial positional relationship of each visual body part by the distribution order of the end of the object. At the same time, the spine-like axis (SPA) was introduced to impose stricter constraints on the spatial structure of the object (the strength of the constraint could be controlled by parameters). This method could retain the main structural information of the object as well as the contour information.

\begin{figure}[t]
	\begin{center}
		\hspace{1pt}
		\includegraphics[width=0.65\linewidth]{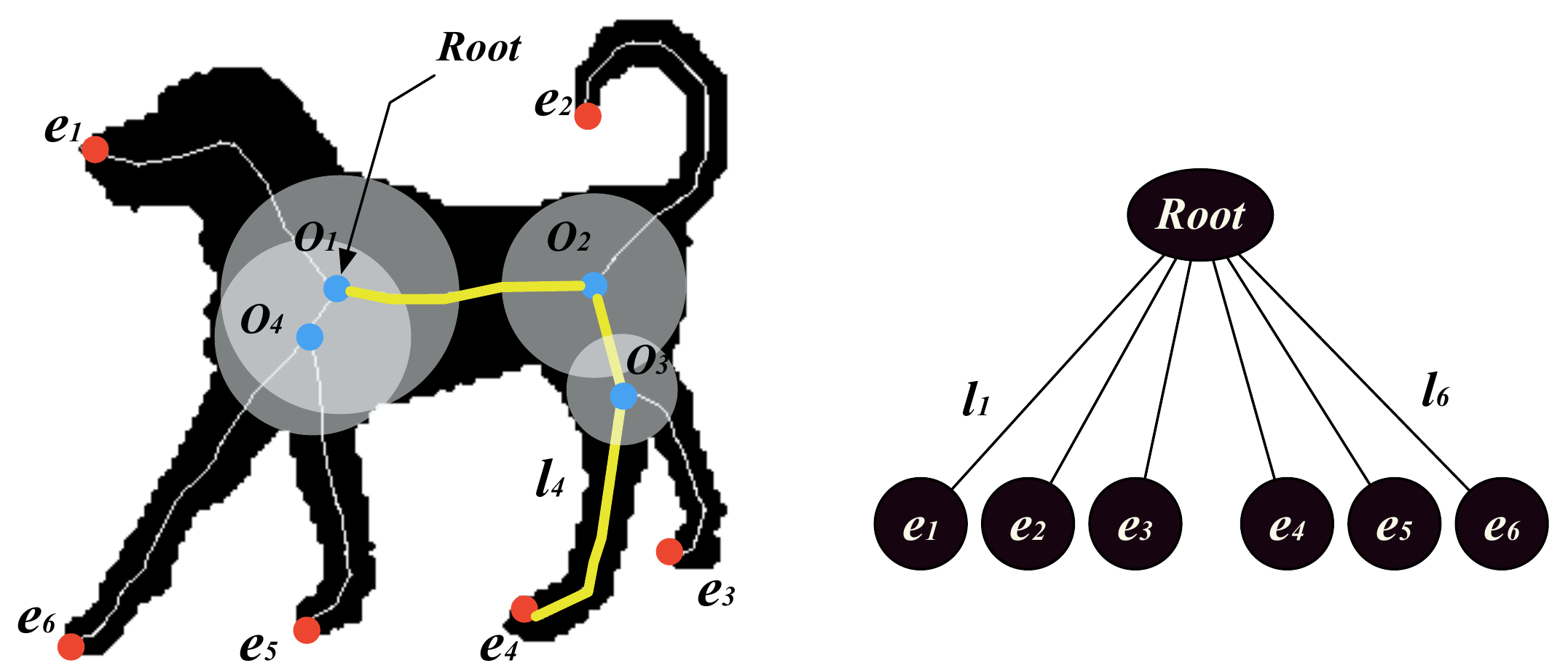}
	\end{center}
	%\vspace{-1.0em}
	\caption{Left: endpoint (red) and junction point (blue) of the skeleton. Right: skeleton tree with height 1.}
	\label{fig:skel}
	\vspace{-.6em}
\end{figure}

%\subsection{骨架}
%\label{sec:motivation}

\begin{figure}[!htb]
	\begin{center}
%		\hspace{1pt}
		\includegraphics[width=0.95\linewidth]{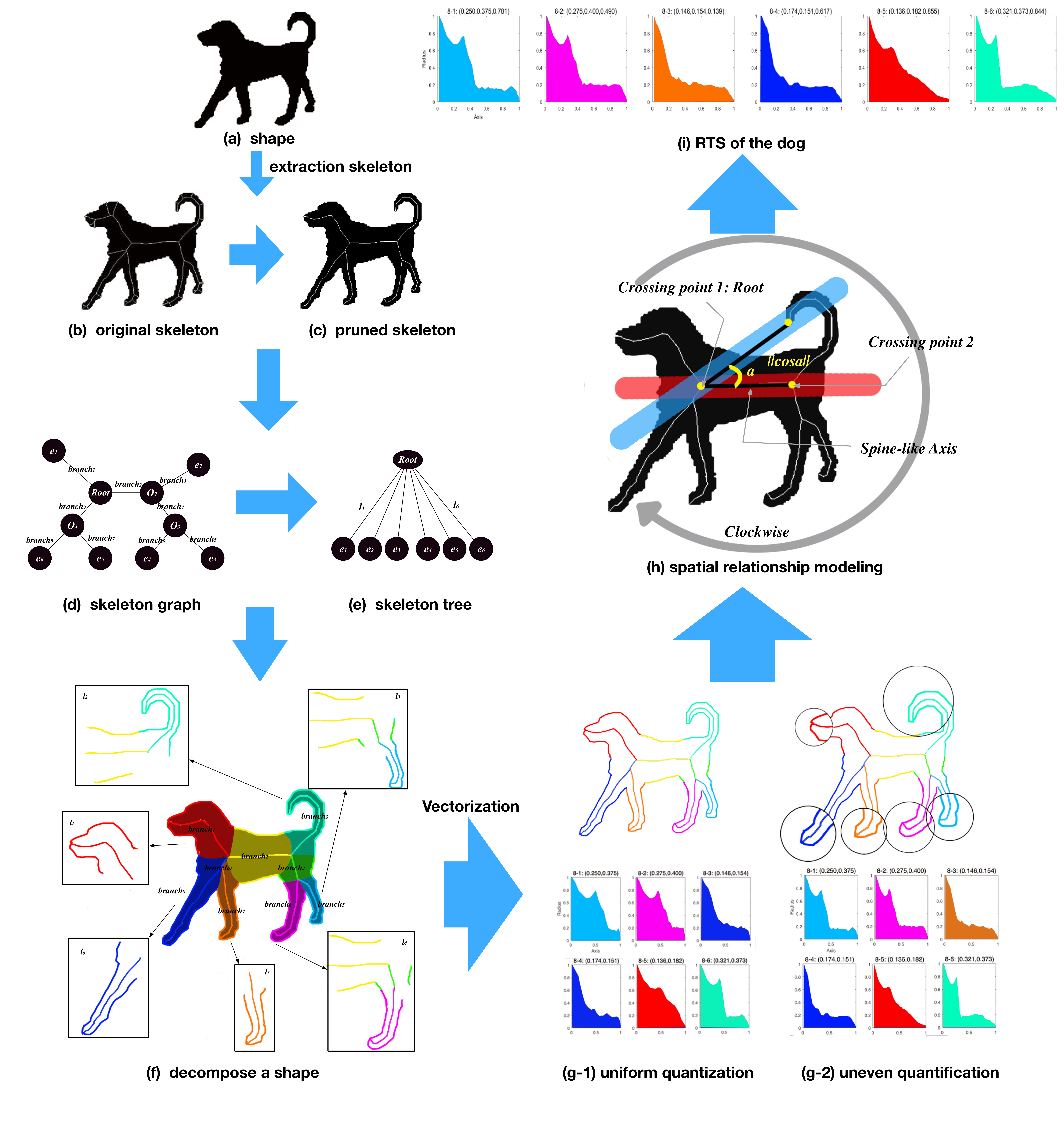}
	\end{center}
	%\vspace{-1.0em} 基于骨架图的物体结构特征的形式化表征过程。
	\caption { The RTS establishment process. (a) The input image. (b) The initial skeleton obtained based on the distance transformation method \cite{borgefors1986distance}. (c) The pruned skeleton obtained based on \cite{Wei2013Skeleton}. (d) Skeleton graph. (e) Considering the stability of the skeleton root node, the skeleton graph is converted into a skeleton tree of height one. (f) Shape decomposition process. (g) Vectorizing the end skeleton path. (h) Model spatial relationships for each end path. The figure shows the counterclockwise spatial constraint weak relationship, and the introduction of the SPA. (We investigate the absolute value of the cosine similarity between the vector consisting of each skeleton end node and the root node and the SPA to make a strong spatial relationship constraint). (i) Forming the final RTS.}
	\label{fig:fc_1}
%	\vspace{-.6em}
\end{figure}

\subsection{Skeleton}
\label{sec:fc_1}

According to the definition of the medial axis by Blum \cite{blum1973biological}, the medial axis or skeleton $S$ is the trajectory (point set) of the center of the largest inscribed circle of a plane $D$. For each skeleton point $s\in{S}$, it stores its maximum circle radius $r(s)$. According to the basic characteristics of the skeleton, here we need to give some relevant definitions about the skeleton.

\textbf{Definition 1}: A skeleton point having only one adjacent point is called an \textbf{endpoint}; a skeleton point having three or more adjacent points is called a \textbf{junction point}. If a skeleton point is neither an endpoint nor a junction point, it is called a \textbf{connection point}. We call the junction point with the largest skeleton radius the \textbf{root point}. For example, as shown in Figure.~\ref{fig:skel}(left), $e_1,e_2,e_3,e_4,e_5,e_6$ are end points while $O_1,O_2,O_3,O_4$ are junction points. Since $O_1$ is the junction point with the largest skeleton radius, it is the root point. All other points are connection points.

\textbf{Definition 2}: A sequence consisting of a connection point between any two connected skeleton points is called a skeleton branch, as shown Figure.~\ref{fig:skel} (left) ($O_1O_2,O_2O_3,O_3e_4$ are skeleton branches). A standard way to build a skeleton graph is as follows: the endpoints and the joint points are the nodes of the graph, and the skeleton branches between all the nodes are chosen as the edges of the graph. For example, Figure.~\ref{fig:fc_1}(d) is a graph representing the skeletons in Figure.~\ref{fig:skel}(left). We try to find a stable representation using the skeleton graph. Since the skeleton joint point is unstable and easily brings about structural changes, and the skeleton root node is relatively stable, we choose to further transform the skeleton graph into a tree with a depth of 1 rooted at the root point, as shown in Figure.~\ref{fig:fc_1}(e). This transformation from graph to tree greatly reduces the complexity of the model’s characterization.

\textbf{Definition 3}: The shortest path from the root node to the end node on the skeleton tree is called a skeleton end path. The endpoint of each skeleton end path corresponds to a visual feature point on the contour of the object. We examine the skeleton path from the root node to the end node to represent the visual part of the endpoint expression. Therefore, based on the root node $R$ and the end node $e_i$, a complete shape can be decomposed into $N$ blocks ($N$ is the total number of end nodes), and each block will be represented by a skeleton end path $l_i: Re_i$, as shown in Figure. ~\ref{fig:fc_1}(f).

\begin{figure*}[t]
	\begin{center}
		\hspace{12pt}
		\includegraphics[width=0.98\linewidth]{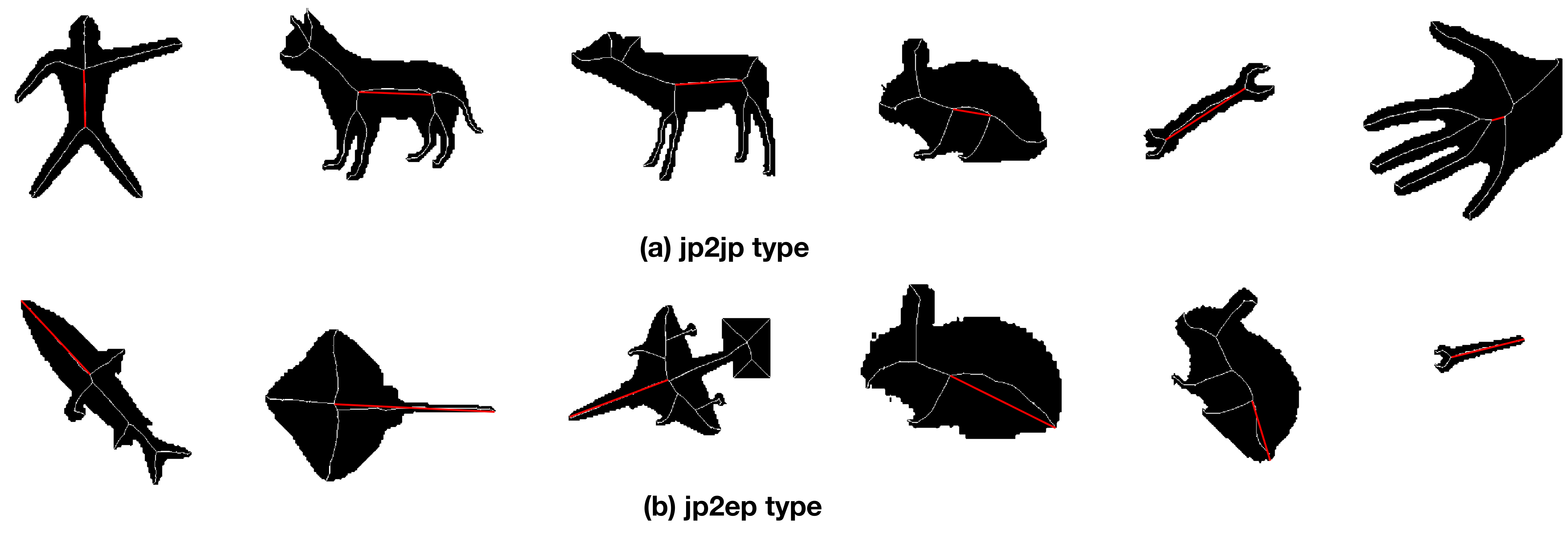}
	\end{center}
	%\vspace{-1.0em}
	\caption{According to the type of the second crossing point, we created two categories of SPA types (the red line is the SPA). (a) $jp2jp-type$. (b) $jp2ep-type$.
	}
%	根据第二穿越点的类型，将spine-like axis的类型分为两大类(红色线标注的为spine-like axis) 。\textbf{top:} $jp2jp$ type spine-like axis. \textbf{down:} $jp2ep$ type spine-like axis.}
	\label{fig:spinetype}
	\vspace{-.6em}
\end{figure*}

\subsection{Constructing Feature Vectors}
\label{sec:quantilization}
Tree-based representation is still a complex method. We only focused on the part from the root node to the end node, which greatly simplified the object's representation. Since the skeleton end path contains both the position information of the skeleton point and the contour information, we quantized the skeleton end path to extract the feature vector. The skeleton path $l_i$ of the root node $R$ to the end node $e_i$ was input into a histogram. The target shape was also described by a histogram, which can fully reflect the skeleton structure and statistical information of the shape. In the statistical histogram, we used the abscissa to represent the path distance of the corresponding root node to the end node, and the ordinate represents the maximum disk radius corresponding to the skeleton point. Because the target shape is diverse, the size is diverse, and the length of the skeleton path is diverse, the original skeleton path must be quantized. We tried two kinds of quantization schemes (both with 50 samples). They are described below.

\textbf{A. Uniform Quantization}

As shown in Figure.~\ref{fig:fc_1}(g-1), in this scheme, uniform spacing sampling was performed for the skeleton branches from the skeleton root node to the skeleton end node. This method is simple and convenient to implement, but treats the mass distribution along the skeleton path equally. This approach ignores the importance of the skeleton path near the end node. Because the end node contains many details of the target shape even though the area ratio of this part is small, it is important for the human eye to distinguish the target shape.

\textbf{B. Uneven Quantification}

We proposed a non-uniform quantization method to quantilize the skeleton path. As shown in Figure.~\ref{fig:fc_1}(g-1), for the end of the skeleton, we uniformly sampled 25 data points (sparse sampling) for the first 80\% of the skeleton paths, and 25 data points (dense sampling) for the last 20\% (end part). This approach introduces the attention mechanism of the human eye, increases the examination of the end portion of the shape

\noindent\textbf{Quantitative Result}

The skeleton path is $\hat{l_i}: \{r_1,r_2,...,r_{50} \} $, where $r_i$ represents the skeleton radius at the $i_{th}$ sampling skeleton point of the skeleton path.
Mass and Length

\noindent\textbf{Mass and Length}

The skeleton path is the most essential feature vector, but we still needed to look for some natural and more representative features. We recorded the skeleton radius of all skeleton points on the skeleton path and the mass $M_i$ as the shape represented by the skeleton path, and measured the length $L_i$ of the skeleton path. Both of them characterize to some extent the importance of the shape represented by the skeleton path as shown in Eq.~\ref{eq:m}.

\begin{equation}\label{eq:m}
M_{i} = \sum_{s\in{l_i}}{r(s)}
\end{equation}

\noindent\textbf{Normalization}

In order to ensure that the RTS has scale invariance, we normalized all the above feature vectors as follows, where $L_{branch_k}$ represents the length of the $k_{th}$ skeleton branch, where all the skeleton branches are mutually exclusive, and the sum of them constitutes a complete skeleton, as shown in Figure.~\ref{fig:fc_1}(f).

\begin{equation}
\begin{split}
\hat{r}(s) &= \frac{r(s)}{r^{*}}, \quad where \ \ r^{*} = \max_{s\in{S}} \ {r(s)}, \\
\hat{M_i} &= \frac{M_i}{M}, \quad where \ \ M = \sum_{s\in{S}}{r(s)}, \\
\hat{L_i} &= \frac{L_i}{L}, \quad where \ \ L = \sum_{branch_k\in{S}}{L_{branch_k}}
\end{split}
\end{equation}

Since only the skeleton path was quantized, the vectors were independent and lacked the constraint of a spatial positional relationship. Therefore, we introduced the constraints of two spatial relationships, weak and strong, to model the spatial topology.

Weak constraint: Although the end node portion of the target object has a relatively large range of motion, there is generally a relatively fixed order of distribution over the contour. Therefore, we looked for the end nodes counterclockwise (or clockwise) from either end node to introduce spatially weak constraints.

Strong constraint: In order to further constrain the distribution of end nodes, we examined the angle of the end nodes relative to the root nodes. Inspired by the fact that almost all objects have relatively stable axes, we introduced a spine-like axis (SPA). The rigid spine-like axis appears to stabilize the shape distribution. We looked at the SPA from several aspects. The axis must cross the skeleton root node, starting at the root node (the first crossing point must be the root node) and traversing the skeleton point (end node or junction) directly connected to the root node.

As shown in Figure.~\ref{fig:fc_1}(d), for all skeleton branches in the skeleton diagram, if both ends of the skeleton branches are joint points, the skeleton branches are $jp2jp-type$; otherwise, they are $jp2ep-type$.

Type 1: $jp2jp-type$. The following conditions should be met:

\begin{itemize}
\item It should have a large proportion of quality and length.
\begin{equation} \label{eq:ml}
jp2jp_{ml} = \alpha \times \hat{M} + (1 - \alpha) \times \hat{L}
\end{equation}

\item It should have a large mass density distribution.
\begin{equation}
jp2jp_{dense} = \frac{\hat{M}}{\hat{L}}
\end{equation}

\item It should have a relatively smooth mass distribution.
\begin{equation}
\begin{split}
jp2jp_{smooth} &= std(l) \\
&= \sqrt{\frac{1}{n}{\sum_{k=1}^n(\hat{r}_i-\bar{r})^2}}
%	\sigma=\sqrt{\frac{1}{n}{\sum_{k=1}^n
%			(x_i-\bar{x})^2}}
\end{split}
\end{equation}
\end{itemize}

Type 2: $jp2ep-type$. One of the endpoints of the skeleton branch is an end point, and the following conditions should be met:

\begin{itemize}
\item It should have a large proportion of quality and length, the same as that expressed in formula.~\ref{eq:ml}.

\item For the skeleton branch from the root node to the end node, the radius value is gradually reduced, and therefore the mass density and mass distribution stability should not be considered. In order to balance the effects of these two missing terms, the following is introduced as a constant term.
\begin{equation}
jp2ep_{cost} = min(jp2jp_{dense}) + std(jp2jp_{dense})
\end{equation}
\end{itemize}

For each skeleton $branch$ in the skeleton graph, a comprehensive evaluation is performed (the constant 10 on the right side of the formula is to balance the order of the two items):

\noindent for $jp2jp-type \ branch$:
\begin{equation}
jp2jp = \beta \times jp2jp_{dense} + (1 - \beta) \times 10 \times jp2jp_{ml}
\end{equation}

\noindent for $jp2ep-type \ branch$:
\begin{equation}
jp2ep = \beta \times jp2ep_{cost} + (1 - \beta) \times 10 \times jp2ep_{ml} 
\end{equation}

It is worth noting that the parameters $\alpha$ and $\beta$ and the $threshold$ are fixed here, and are fixed for all experiments: $\alpha=0.65,\beta=0.3, and threshold=0.224$.

After finding the SPA: $\overrightarrow{Axis}$, in order to constrain the spatial distribution of the end nodes, we used the absolute value of the cosine similarity between $\overrightarrow{Vec_i}$ (the vector of the skeleton root node $R$ to the skeleton end node $e_i$) and $\overrightarrow{Axis}$ (the direction of Axis is not constrained, so absolute values are added) to represent the spatial distribution of the end node relative to the SPA.

\begin{equation}
	Vi = \left \| \frac{\overrightarrow{Axis} \cdot \overrightarrow{Vec_i} }{\left \| \overrightarrow{Vec_i} \right \| \left \| \overrightarrow{Vec_i} \right \|} \right \|
\end{equation}

\vspace{1.2em}
\noindent\textbf{Algorithm Description}
%\noindent\textbf{算法描述}
\makeatletter
\def\BState{\State\hskip-\ALG@thistlm}
\makeatother
\begin{algorithm*}[htp]

	\caption{Algorithm to find the 'Spine-like Axis'}\label{euclid}
	\label{alg:spine}
	\begin{algorithmic}[1]
		\Procedure{find-spineAxis}{$S,\beta,threshold$} %$S,l_i,EP,JP,\beta,threshold$
		\State $S$ is the pruned skeleton of a shape.
		\State Let $branch_i:(\hat{r}_1,\hat{r}_2,...,\hat{r}_N)$ be the skeleton branch $i$. $N$ is the number of skeleton points in the branch.
		\State Let $M_i$ be the mass of $branch_i$ and $L_i$ be the length of $branch_i$.
%		\State Let $\hat{l_i}:(\hat{r_1},\hat{r_2},...,\hat{r_{50}})$ be the quantification result of $l_i$.
		\State Let $EP$ be the set of end points of $S$ and $JP$ be the set of junction points. $R$ is the root point of $S$. % $P = EP \cup JP$.
		\For{each $branch_i \in S$}
			\If{$branch_i \in jp2jp type$}
				\State $jp2jp_{ml_i} \gets \alpha_1 \times{\hat{M}_i} + (1 - \alpha_1) \times{\hat{L}_i}$
				\State $jp2jp_{dense_i} \gets \frac{\hat{M}_i}{\hat{L}_i}$
				\State $jp2jp_{smooth_i} \gets std(branch_i)$
			\Else
				\State $jp2ep_{ml_i} \gets \alpha_1 \times{\hat{M}_i} + (1 - \alpha_1) \times{\hat{L}_i}$
			\EndIf
		\EndFor
		\State $jp2ep_{cost} \gets min(jp2jp_{ml}) + std(jp2jp_{ml})$
%		\For{each $p_i \in P$}
%			\If{}
%		\EndFor
		\State $jp2jp_i \gets \beta \times jp2jp_{dense_i} + (1 - \beta) \times 10 \times jp2jp_{ml_i}$ 
		\State $jp2ep_i \gets \beta \times jp2ep_{cost} + (1 - \beta) \times 10 \times jp2ep_{ml_i} $
		\State $JP2P \gets jp2jp \cup jp2ep$
% 		\State $\textit{stringlen} \gets \text{length of }\textit{string}$
%		\State $i \gets \textit{patlen}$
		\State Turn $S$ to undirected graph $G(V,E)$, $V$ is the set of points, $V$ is $EP \cup JP$, $E$ is the comprehensive value ($JP2P$) of $branch$.
		\State Find all points $P$ link to $R$ directly.
		\State Initialize $crossingpoint_2 \gets \mathop{\arg\max}_{p_i \in P}{JP2JP} $
		\While{$crossingpoint_2 \in JP \land jp2jp_{smooth_{crossingpoint_2}} > threshold$}
			\State Delete the value of this point in $JP2P$.
			\State Update $crossingpoint_2 \gets \mathop{\arg\max}_{p_i \in P}{JP2P}$
		\EndWhile
		\State $crossingpoint_1 \gets R$
		\State \Return{$crossingpoint_1 - crossingpoint_2$}
%		\For{each $p_i \in P$}
%			\State if $p_i \in JP$
%		\EndFor
%		\BState \emph{top}:
%		\If {$i > \textit{stringlen}$} \Return false
%		\EndIf
%		\State $j \gets \textit{patlen}$
%		\BState \emph{loop}:
%		\If {$\textit{string}(i) = \textit{path}(j)$}
%		\State $j \gets j-1$.
%		\State $i \gets i-1$.
%		\State \textbf{goto} \emph{loop}.
%		\State \textbf{close};
%		\EndIf
%		\State $i \gets i+\max(\textit{delta}_1(\textit{string}(i)),\textit{delta}_2(j))$.
%		\State \textbf{goto} \emph{top}.
		\EndProcedure
	\end{algorithmic}
\end{algorithm*}

\begin{figure}[t]
	\begin{center}
		\hspace{12pt}
		\includegraphics[width=0.75\linewidth]{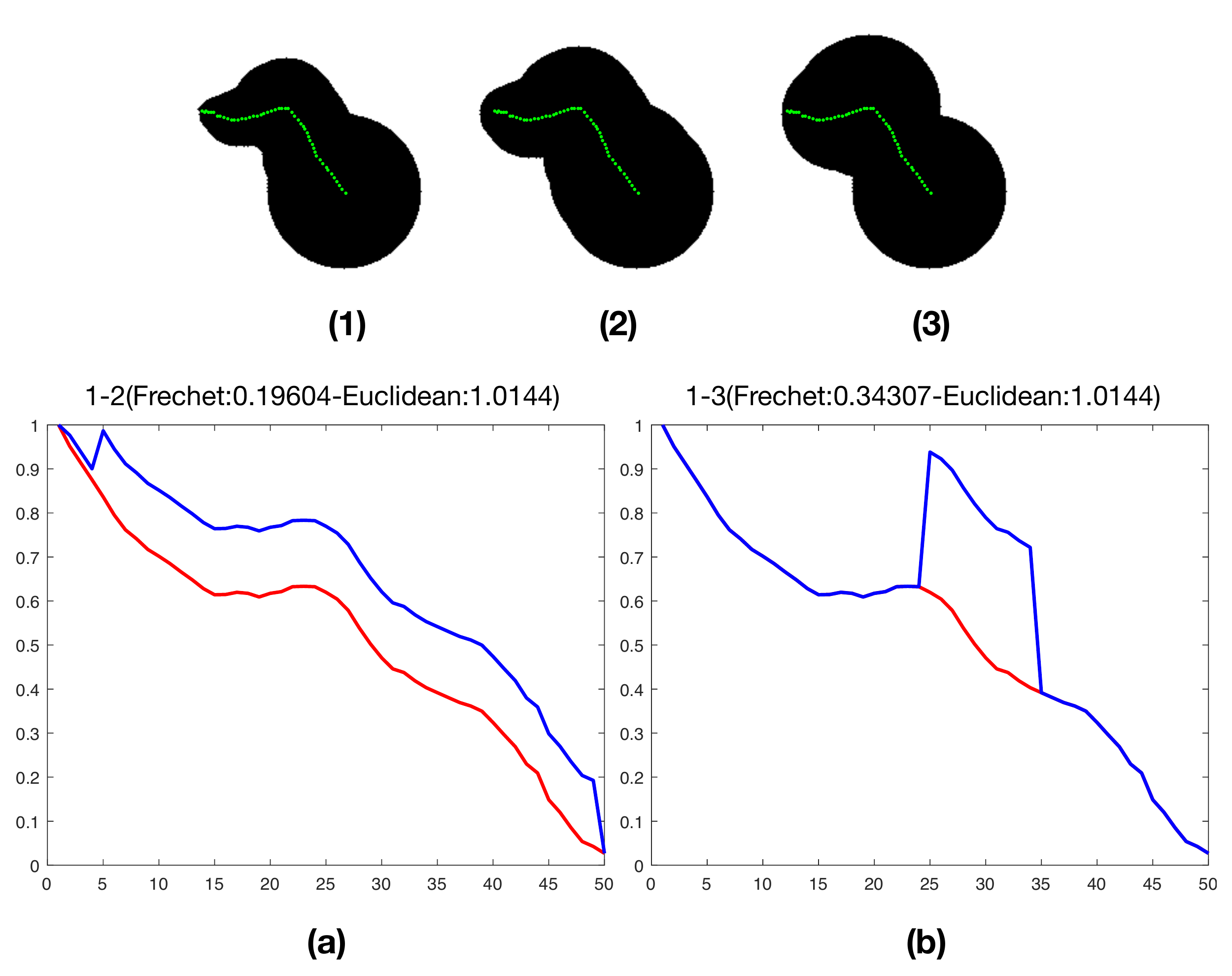}
	\end{center}
	\vspace{1.0em}
	\caption{ Comparison of two distance functions. The green dots in figures (1), (2), and (3) are skeleton points. (a) Distribution of (1) and (2) skeleton paths, which are almost completely parallel. The two skeleton paths have a Fr{\'e}chet distance of only 0.19604 and an Euclidean distance of 1.0144. (b) Distribution of (1) and (3) skeleton paths. The Euclidean distance of the two skeleton paths does not change, and the Fr{\'e}chet distance increases significantly from 0.19604 to 0.34307. From our visual observation of the three shapes, it is obvious that (1) and (2) are similar in shape, and (1) and (3) have a large difference in shape. Therefore, from the results of the two distance measures, we conclude that the Fr{\'e}chet distance can better reflect this difference, which is more in line with human visual cognition.
	}
	\label{fig:dis_cmp}
	\vspace{-.6em}
\end{figure}

\begin{figure*}[t]
	\begin{center}
		\hspace{12pt}
		\includegraphics[width=0.95\linewidth]{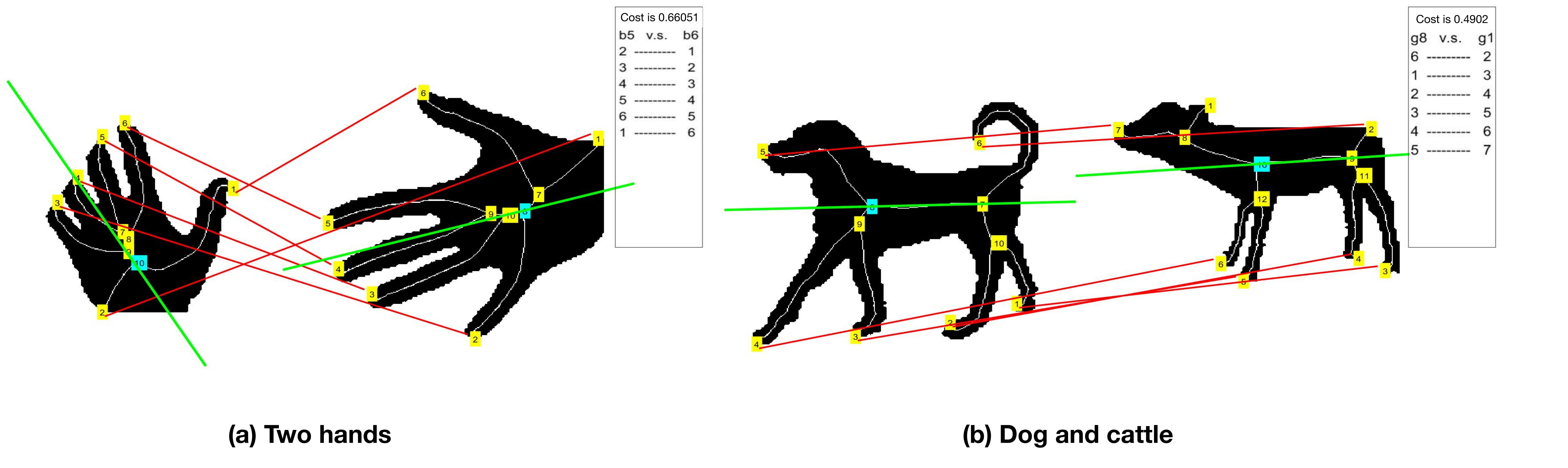}
	\end{center}
	%\vspace{-1.0em}
	\caption{The corresponding end nodes between the two skeletons are linked with red lines. The green line represents the SPA obtained by Algorithm.~\ref{alg:spine} presented in Section.~\ref{sec:quantilization}. The matching cost (distance or non-similarity) of the two shapes and the list of endpoint correspondences are shown in the legend in the upper right corner.
%	The corresponding end nodes between the two skeletons are linked with red lines. The green line in the figure represents the SPA obtained by Algorithm.~\ref{alg:spine} in Sec.~\ref{sec:quantilization} . The matching cost (distance or non-similarity) of the two shapes and the list of endpoint correspondences are shown in the block diagram in the upper right corner.
	}
%	The corresponding end nodes between the two skeletons are linked with red lines. 图中的绿线表示用Section.~\ref{sec:quantilization}中的Algorithm.~\ref{alg:spine}得到的'spine-like axis'。右上角的框图中展示了两个形状的匹配代价(距离或称非相似度)以及端点对应关系列表。}
	\label{fig:match}
	\vspace{-.6em}
\end{figure*}

For a specific shape $\mathcal{X}$, the complete process of the RTS is shown in Figure.~\ref{fig:fc_1}. 
$\mathcal{X}$ has $N$ end nodes, they form a sequence according to the clockwise ordering of the contours: $\left\{E_1,E_2,\cdots,E_N\right\}$, and the final characterization results are expressed as: 
\[E_i = \left\{ \hat{l}_i:\left\{ \hat{r}_1,\hat{r}_2,\cdots,\hat{r}_{50}
\right\},\hat{M}_i,\hat{L}_i,V_i
\right\}_{i=1}^{N}
\]

\section{Similarity Measure}
\label{sec:simi}
Based on the RTS characterization, the skeleton end nodes of each target shape can be measured by three major parts: the quantized vector of the skeleton path $\hat{l}:\left\{ \hat{r}_1,\hat{r}_2,...,\hat{r}_{50} \right \}$; the normalized mass $\hat{M}$ and length $\hat{L}$; and the spatial distribution $V$ relative to the SPA.

\subsection{Improved Distance Metrics}
In the past, the similarity of the skeleton path vector was examined by Euclidean distance, but this method is not always ideal. What we really want to look at is the trend of the path distribution, not the sum of the differences. As shown in Figure.~\ref{fig:dis_cmp}, the distributions of the two skeleton paths in (a) are completely consistent, but due to the difference in absolute values, the difference between the two is larger if measured by the Euclidean distance. However, the distributions of the two skeleton paths in (b) are quite different, yet the value of the Euclidean distance metric is exactly the same as in (a). Moreover, the difference in shape cannot be detected, which is contrary to visual cognition. This increase in the value of the Fr{\'e}chet distance results in this difference in shape. Therefore, we proposed a reconsideration of the distance. By using the Fr{\'e}chet distance instead of the Euclidean distance, we can well examine the similarity of the spatial distribution of the skeleton path. Since we have quantized the skeleton path in Section.~\ref{sec:quantilization}, we used the $discrete \ Fr $\'{e}$chet \ distance$ \cite{eiter1994computing}.

Let $S$ and $S'$ denote two skeletons to be matched, and let $u$ and $v$ be the skeleton radius of $S$ and $S'$, respectively. Let $P$ and $Q$ be the skeleton end paths in $S$ and $S'$, respectively. We obtain $P:(u_1,u_2,...,u_{50})$ in $S$ and $Q:(v_1,v_2,...,v_{50})$ in $S'$. Then we compute the path distances between the two end paths $P$ and $Q$. A coupling $L$ between $P$ and $Q$ is a sequence
\[
	(u_{a_1},v_{b_1}),(u_{a_2},v_{b_2}),...,(u_{a_m},v_{b_m})
\]
of distinct pairs from $P \times Q$ such that $a_1 = b_1=1$, $a_m = b_m = 50$, and for all $i=1,...,50$ we have $a_{i+1} = a_i$ or $a_{i+1} = a_i + 1$, and $b_{i+1}=b_i$ or $b_{i+1}=b_i$. Thus, a coupling $L$ is the length of the longest link in $L$, that is (here we can choose the Euclidean distance),
\begin{equation}
	\left\| L \right\| = \max_{i=1,..,m}{d(u_{a_i},v_{b_i})}
\end{equation}

The $discrete \ Fr $\'{e}$chet \ distance$ between $P$ and $Q$ is
\begin{equation}
	d_F(P,Q) = \min{\left\| L\right\| }
\end{equation}

\subsection{Comprehensive Similarity Measure}
Based on the RTS representation of the target shape we proposed in Section.~\ref{sec:shape_res}, we examined the distance metrics of the two skeleton end paths from three aspects:

\begin{itemize}
\item For the similarity measure of the skeleton path vector, we replaced the original Euclidean distance with the Fr{\'e}chet distance, and better examined the similarity of the path in the spatial distribution.
\begin{equation}
	d_F = Fr\acute{e}chet(P,Q)
\end{equation}

\item Comprehensive examination of mass and length: We introduced the parameter $\alpha$ to reconcile the two values of mass difference and length difference. It is worth noting that this $\alpha$ and the parameter $\alpha$ in formula.~\ref{eq:ml} in Section.~\ref{sec:quantilization} are the same value (0.65).

\begin{equation}
\centering
\begin{split}
	d_{m} = \frac{(M_p - M_q)^2}{M_p+M_q} \\
	d_{l} = \frac{(L_p - L_q)^2}{L_p+L_q} \\
	d_{ML} = \alpha \times d_{l} + (1-\alpha )\times d_m\\ 
\end{split}
\end{equation}

\item For the similarity of the distribution of end nodes relative to the root node, we measured the difference of the square of the difference between the values of the spatial distribution between the two end nodes.

\begin{equation}
	d_V = \frac{(V_p - V_q)^2}{V_p+V_q}
\end{equation}
Comprehensive considerations:
%综合考量:
\begin{equation}
\label{con:dis}
	d(P,Q) = d_F + \beta_1 \times d_{ML} + \beta_2 \times d_V
\end{equation}

\end{itemize}

\noindent\textbf{Comprehensive Considerations:}

It is worth noting that the only parameters that can be adjusted for the whole process are $\beta_1$ and $\beta_2$, where $\beta_2$ is the adjustment of the constraint strength of the spatial result. For rigid objects, the distribution is more regular, and the value of $\beta_2$ can be larger. For non-rigid objects, the spatial distribution of the structure is relatively weak, so the value of $\beta_2$ can be small. The range of $\beta_2$ is generally 0$-$1.

\subsection{Shape Matching}
We examined the matching relationship between the two shapes and the distance metric. According to spatial constraint 1 mentioned in Section.~\ref{sec:quantilization}, the distribution order of the skeleton on the contour is expressed as a sequence. We used the OSB algorithm \cite{latecki2007optimal} to examine the matching relationship between the two shapes. The OSB algorithm can handle the matching problem between two unequal length sequences. Since the sequence may contain some abnormal elements, it is necessary to skip some elements in the two sequences that do not match, thus achieving elastic matching. However, skipping too many elements increases the likelihood of an unexpected match. To avoid this, the OSB algorithm adds a certain penalty to skipping elements. Let $S$ and $S'$ be two skeletons to be matched. For the end nodes in $S$ and $S'$, the sequences are sequenced clockwise to form sequence $\left( P_1,P_2,...,P_M\right)$ and sequence $\left( Q_1,Q_2,...,Q_N\right)$, where $M$, and $N$, are the number of end nodes in the two skeletons, respectively ($M<N$). According to our distance measurement method above, the distance matrix $D$ is formed between the two sequences:

\begin{equation}
	D(S,S') = \begin{pmatrix}
d(P_1,Q_1) & d(P_1,Q_2) & \cdots & d(P_1,Q_N) \\ 
d(P_2,Q_1) & d(P_2,Q_2) & \cdots & d(P_2,Q_N)\\
\vdots & \vdots & \ddots & \vdots \\
d(P_M,Q_1) & d(P_M,Q_2) & \cdots & d(P_M,Q_N)\\
	\end{pmatrix}
%	D(S,S') = \left ( \begin{array}{ccc}
%d(P_1,Q_1) & d(P_1,Q_2) & \cdots & d(P_1,Q_N) \\ 
%d(P_2,Q_1) & d(P_2，Q_2) & \cdots & d(P_2,Q_N)\\
%\vdots & \vdots &
%\end{array}
%\right)
\end{equation}

Our goal becomes finding an optimal matching method so that the total cost of matching is the smallest. In this paper, $d(P_i,Q_{\infty})=jumpcost$ is computed as follows \cite{bai2008path}:

\begin{small}
	\begin{equation}
	jumpcost=mean_i(min_j(d(P_i,Q_j))) + std_i(min_j(d(P_i,Q_j))).
	\end{equation}
\end{small}

The directed matrix was constructed with a directed acyclic graph (DAG). We can find the optimal match by using the shortest path algorithm on the DAG. The nodes of the DAG are all index pairs $(i,j) \in \left\{ 1,\cdots,M\right\} \times \left\{1,\cdots,N\right\} $, and the edge weights $w$ are defined as:

\begin{small}
	\begin{equation}
	\!w((i,j),(k,l))\!\! =\!\!\!\! \\
	\left \{
	\begin{aligned}
	&d(P_i,Q_j)& & {if \!\ k\!-\!i\!=\!1 \ and \ l\!-\!j\!=\!1 }\\
	&\!(k\!-\!i\!-\!1\!)\! \cdot{\!jumpcost}\!\!\!\!\!& & {if\! \ k\!-\!i\! >\! 1 \ and \ l\! -\! j\! >\!1}\\
	&\infty & &otherwise\\
	\end{aligned}
	\right.
	\end{equation}
\end{small}
 \begin{equation}
\centering
\begin{split}
(cost, f) = OSB(D(S,S')), \quad f \ is \ a \ correspondence
\end{split}
\end{equation}

The cost here is the matching $cost$ (non-similarity or distance) between the RTSs, and $f$ is the matching correspondence between the RTSs.

We give two simple examples illustrating our matching approach. The left of the part of Figure.~\ref{fig:match} shows the end node matching relationship between the skeletons of two different hands. The two hands not only have a large rotation angle, but also have occlusion deformation. However, we still achieved a perfect match. This is not only because of our good similarity measurement algorithm, but also due to the important role of the spine-like axis. Effectively dealing with rotation invariance thanks to our proposed spine-like axis, we will conduct comparative experiments in Section.~\ref{sec:exprecog} to emphasize the superiority of this approach. The right part of Figure.~\ref{fig:match} shows skeletons of a dog and a cow with the corresponding end nodes linked by lines. We have marked the endpoints that match each other, although node 1 in the cattle has no corresponding end node. This again confirms the performance of the elastic matching of the OSB algorithm, which can skip some of the abnormal end nodes in the matched skeleton map. This is very suitable for the case of local shape matching.

However, we do not just need to get the correspondence between the endpoints, more importantly, we need to get the global similarity. What is obtained only by the above method is the similarity result of the local matching. This approach ignores the matching cost of the unequal number of nodes on both ends. As shown in the figure, because the rabbit has few endpoints, its matching cost with the cat is very low, and the matching value of the cat is much lower than that of other more similar mammals (such as dogs). This similarity can no longer evaluate the global similarity of the two quadrupeds. Therefore, we further penalized the difference in the number of endpoints. Our approach combined the advantages of the OSB algorithm in finding matching correspondences while ensuring the overall matching cost, and is as follows:

\begin{equation}
\centering
\begin{split}
const = \frac{cost}{ \min(M,N)}\\ %, \quad M = \min(M,N) \\
cost = cost + const * |M-N| \\
\end{split}
\end{equation}

\begin{figure}[!htb]
	\begin{center}
%		\hspace{12pt}
		\includegraphics[width=0.77\linewidth]{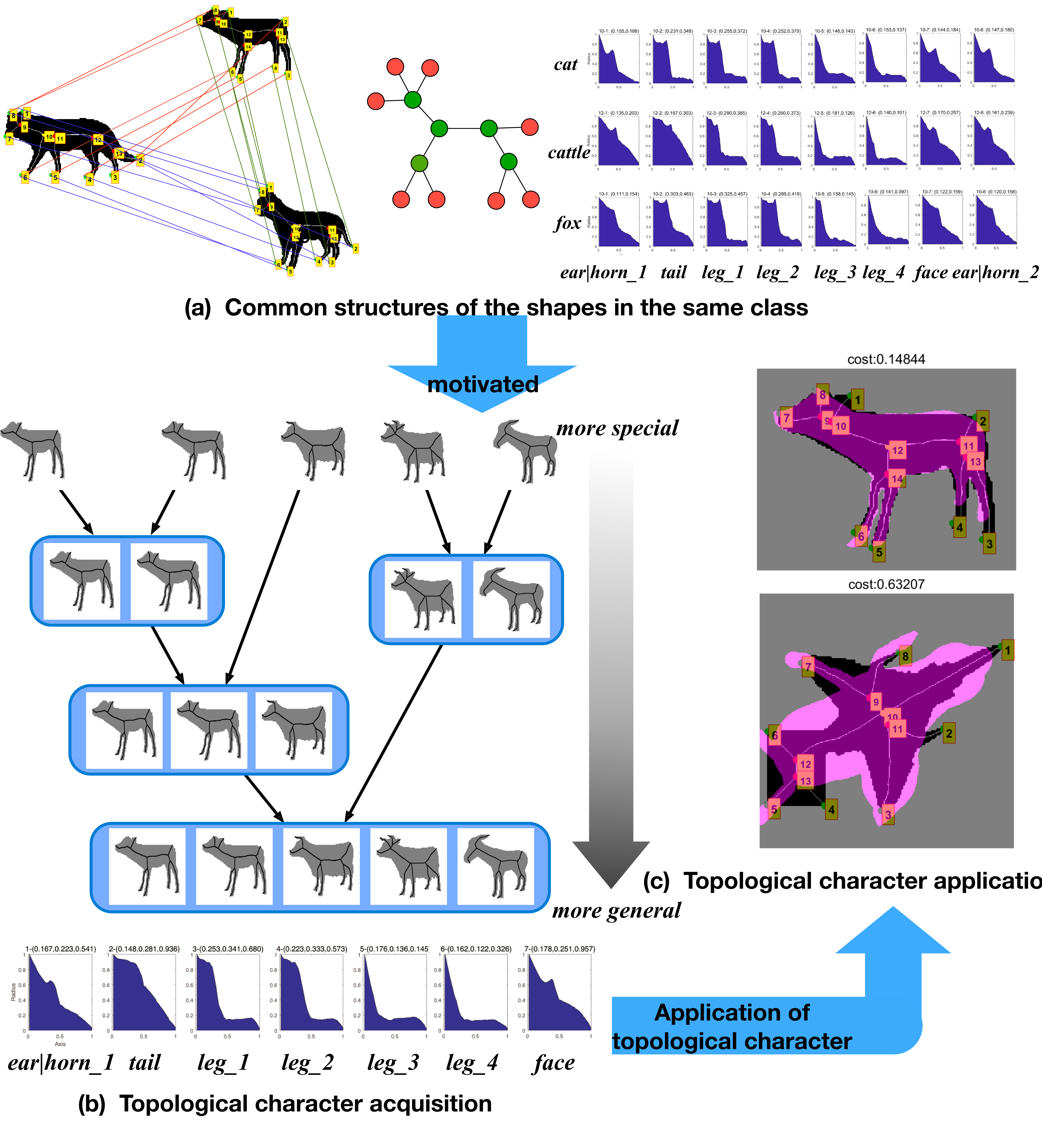}
	\end{center}
	%\vspace{-1.0em}
	\caption{The process of topological character acquisition of inductive learning based on multi-level generalization. (a) Common structures of shapes in the same category. Left: A fox, cat, and cattle with their corresponding points linked by lines. Middle: Common skeleton graphs. Right: Similar distribution of the skeleton paths. (b) The process of multi-level generalization. First, five shapes were initialized as five clusters. In every step, the two most similar clusters were generalized into a new cluster. Finally, the five shapes were generalized into one cluster. Thus, we obtained the characterization of the common structure of the genus (GRTS), which was called topological character acquisition. (c) Topological character application. We applied the obtained GRTS to the similar class (top: a cow) and a heterogeneous class (down: a airplane). The purple mask was the result of topological character application. When we applied the topological character to similar instances, the matching value is very low. while, places where the mask did not cover indicated differences. When the topological character was applied to a heterogeneous class, the matching generation value was high. Here the meaning of the mask was not so important, but it could be seen that the mask attempted to extend outward to characterize the limbs in the four protruding parts of the airplane.
%	 In summary, we could see that our topological character can not only correctly identify categories, but also has strong interpretability and can explore the specific differences between instances and categories.
}

	\label{fig:fc_2}
%	\vspace{-.6em}
\end{figure}

\section{Multi-level Generalization for Inductive Learning}
We proposed a multi-level structure generalization framework based on the RTS, carried out inductive learning of 'seeking common ground', and gradually explored the common structure of the same kind of shape. This common structure contains the intrinsic structural information of a class of objects, and achieves a generalization of the common features of the generics. Our motivation is that similar objects have a similar structure, and this similar structure is reflected in the RTS it establishes, as shown in Figure.~\ref{fig:fc_2}(a).

Based on the shape similarity measure algorithm presented in Section.~\ref{sec:simi}, we first initialized each shape in the generic class as a special case, and then iteratively generalized the two most similar objects into a new more general instance until the number of instances as reduced to 1. The generalization is a process from specific to general, also called inductive-learning, while the information of the same shape can be extracted from the current instance to provide a more general representation for subsequent merge iterations. It is worth noting that although the above inductive learning process is a bit like clustering, the two are completely different. Inductive learning is semi-supervised, with the aim of obtaining an explicit, generalized formal representation. The result is declarative knowledge; conversely, the clustering is unsupervised, aiming at minimizing the distance within the category and maximizing the distance between classes. The result was the class’s center position.

\subsection{Establish the Generalized RTS (GRTS)}
The GRTS (generalized representation of topological structure) is built from a framework of hierarchical generalization. Each skeleton end path in the GRTS consists of a cross-correlation end-of-branch path that belongs to all RTSs. A very important principle must be followed while obtaining the GRTS: the shape voting principle.

\begin{figure}[!htb]
	\begin{center}
		\hspace{12pt}
		\includegraphics[width=0.45\linewidth]{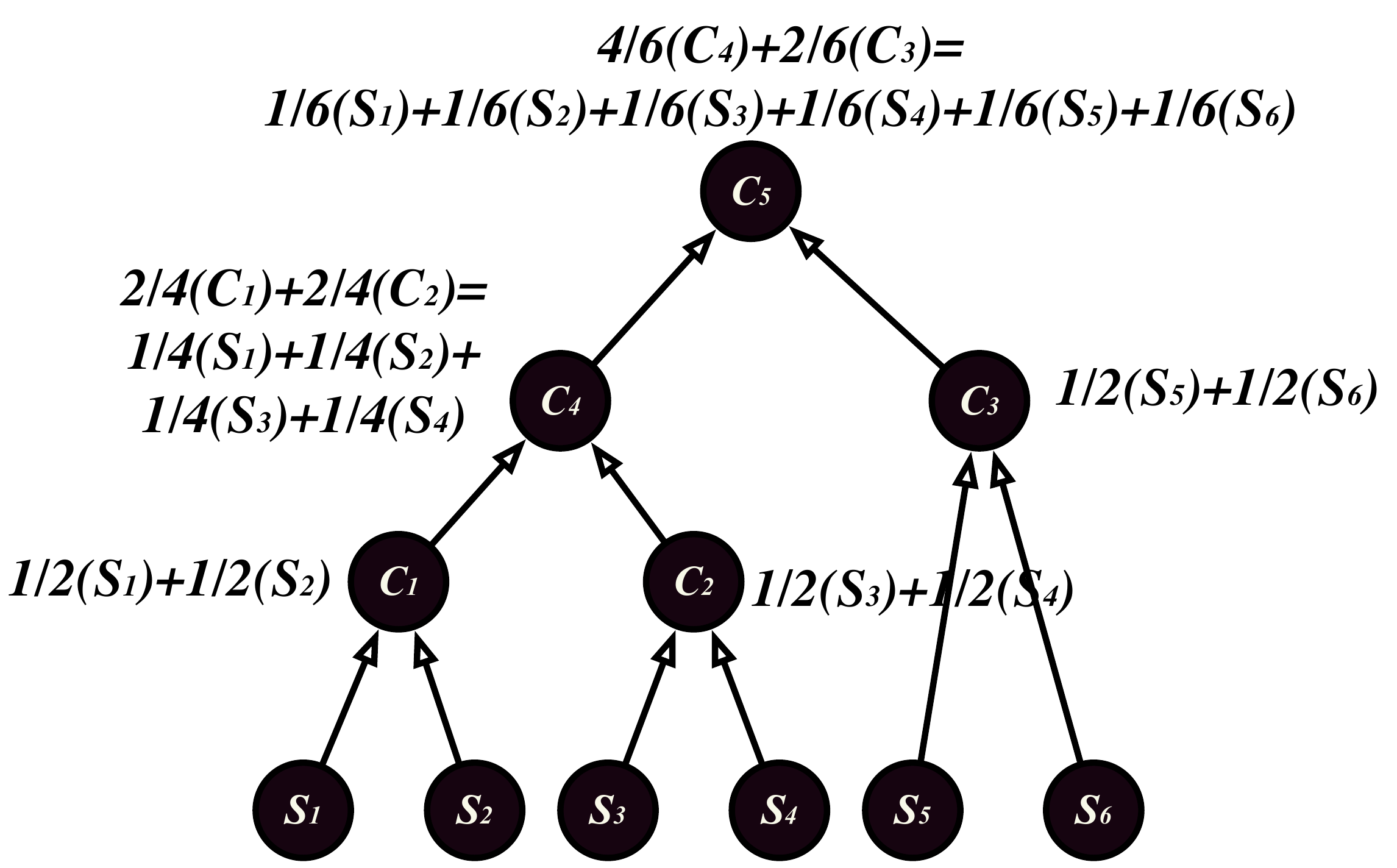}
	\end{center}
	\vspace{-1.0em}
	\caption{The fairness of the shape voting principle. We can see the fairness of this approach by weighing the number of instances contained in each node. Each node is fairly weighted by the instances it contains. For example, node C5 contains the instances $\left\{S_1, S_2, S_3, S_4, S_5, S_6\right\}$ and merged them. Therefore, each shape will have the same effect on the final structure of $C_5$: $C_5=\frac{1}{6}S_1 + \frac{1}{6}S_2 +\frac{1}{6} S_3+\frac{1}{6}S_4+\frac{1}{6}S_5+\frac{1}{6}S_6$.}
	
	\label{fig:fair}
	\vspace{-0.6em}
\end{figure}

\noindent\textbf{Shape Voting Principle}

The minority obeys the majority. The number of instances belonging to the GRTS is a very important piece of information: a node consisting of very few instances is more likely to be an exception node. A node consisting of a large number of instances should be an important node.What we want to extract is a general representation of a common structure. This process can be seen as group decision-making. Shapes of the same genus that have an absolute quantitative advantage play a decisive role in our perception of such objects. We often use heterogeneous to express the few shapes that are unusual. Therefore, in the process of generalizing GRTS, we should give more weight to those examples with quantitative advantages. A natural practice is to assign different weights to different nodes in the GRTS depending on the number of instances. Let a GRTS have $\lambda$ shapes, and the generalization to get one of its nodes is generalized with $\mu$ shapes. Then the node has a weight of $\omega=\frac{\mu}{\lambda}$. We can do a simple calculation to find the fairness of this approach, that is, the principle of one vote per person, as shown in Figure.~\ref{fig:fair}. From this we can derive a more general summary structural representation of the same genus. 

The $GRTS_x$ node is generalized by $\upsilon$ instances, and the $GRTS_y$ node is generalized by $\mu$ instances. Let $GRTS _x$ contain $N$ end nodes: $E_x=\left\{e_{x_1},\cdots,e_{x_N}\right\}$. $GRTS_y$ contains $M$ end nodes: $E_y=\left\{e_{y_1},\cdots,e_{y_M}\right\}$.

For the generalized results $GRTS$ of $GRTS _x$ and $GRTS _y$, we first examined the shape matching results of GRTSx and GRTSy, and found the correspondence between the skeleton end nodes of the two shapes $\phi:\left\{e_{x_1},\cdots,e_{x_N}\right\} \rightarrow \left\{e_{y_1},\cdots,e_{y_M}\right\}$,
\small\[
	(e_{x_{a_1}},e_{y_{b_1}}),(e_{x_{a_2}},e_{y_{b_2}}),\cdots,(e_{x_{a_m}},e_{y_{b_m}}), \ m \le \min(N,M)
\]

Then, the resulting $GRTS$ has $m$ skeleton end nodes, and each end node $e_i$ is obtained by matching $(e_{x_{a_i}},e_{y_{b_i}})$. 
\vspace{-0.5em}
\[e_{x_{a_i}}:\left\{ l_{x_{a_i}} \rightarrow \left\{r_{x_1},\cdots,r_{x_{50}}\right\},M_{x_{a_i}},L_{x_{a_i}},V_{x_{a_i}}\right\} 
\]
\[e_{y_{b_i}}:\left\{
l_{y_{b_i}}\rightarrow \left\{r_{y_1},\cdots,r_{y_{50}}\right\},M_{y_{b_i}},L_{y_{b_i}},V_{y_{b_i}}\right\} 
\]
\[
e_i:\left\{ l_i \rightarrow \left\{
r_1,\cdots,r_{50}
\right\},M_i,L_i,V_i
\right\}
\]

Then our generalization strategy is as follows:
\vspace{-0.5em}
\begin{eqnarray}
li = \frac{\upsilon}{\upsilon+\mu} \ l_{x_{a_i}} + \frac{\mu}{\upsilon+\mu} \ l_{y_{b_i}}, \\
M_i = \frac{\upsilon}{\upsilon+\mu} \ M_{x_{a_i}} + \frac{\mu}{\upsilon+\mu} \ M_{y_{b_i}}, \\
L_i = \frac{\upsilon}{\upsilon+\mu} \ L_{x_{a_i}} + \frac{\mu}{\upsilon+\mu} \ L_{y_{b_i}} ,\\
V_i = \frac{\upsilon}{\upsilon+\mu} \ V_{x_{a_i}} + \frac{\mu}{\upsilon+\mu} \ V_{y_{b_i}} 
\end{eqnarray}

%\section{Front matter}
%
%The author names and affiliations could be formatted in two ways:
%\begin{enumerate}[(1)]
%\item Group the authors per affiliation.
%\item Use footnotes to indicate the affiliations.
%\end{enumerate}
%See the front matter of this document for examples. You are recommended to conform your choice to the journal you are submitting to.
%
%\section{Bibliography styles}
%
%There are various bibliography styles available. You can select the style of your choice in the preamble of this document. These styles are Elsevier styles based on standard styles like Harvard and Vancouver. Please use Bib\TeX\ to generate your bibliography and include DOIs whenever available.
%
%Here are two sample references: \cite{Feynman1963118,Dirac1953888}.

\subsection{Distance Metric Between GRTSs}
Since we kept the global object structure representation unchanged, we still chose the distance metric presented in Section.~\ref{sec:simi}. Let $GRTS_x$ and $GRTS_y$ denote two nodes to be generalized. Let $d(e_{x_i},e_{y_j})$ denote the distance between two end points, where $(i,j) \in \left\{ 1,\cdots,N\right\}\times\left\{1,\cdots,M\right\}$. We obtained a distance matrix computed with the following formula:
\vspace{-1.0em}
\begin{equation} \label{con:csdm}
	D(GRTS_x,GRTS_y) = \begin{pmatrix}
	d(e_{x_1},x_{y_1}) & \cdots & d(e_{x_1},e_{y_M}) \\
	\vdots &\ddots &\vdots \\
	d(e_{x_N},x_{y_1}) & \cdots & d(e_{x_N},e_{y_M}) \\
	\end{pmatrix}
\end{equation}

By applying the OSB algorithm to the matrix in formula.~\ref{con:csdm}, we obtained the dissimilarity and correspondence between $GRTS_x$ and $GRTS_y$.
\begin{equation}
	(cost,f) = OSB(D(GRTS_x,GRTS_y))
\end{equation}
\vspace{-0.2em}
The $cost$ here is the matching cost (non-similarity or distance) between the obtained GRTSs, and $f$ is the matching relationship between the GRTSs.

\subsection{Generalization Scheme}
We generalized the same type of objects layer-by-layer using the generalization scheme, and at the same time got the common structure representation while the two GRTSs were generalized until all instances were generalized into a single node. The final GRTS was our general representation of the general structure of this generic class. Now we demonstrate our generalization scheme. We are given some labeled shape set $X=\left\{S_1,S_2,\cdots,S_N \right\}$, which belongs to the same category and is to be merged into one cluster $\mathcal{C}=\left\{ GRTSs \right\}$, where $N$ is the number of shapes. The representation of each shape can be seen as a definition of the shape. According to the information in the training example, we generalized set $H$ and gradually reduced it. Eventually, $H$ converged to contain only the required rules, and the final rule was the highest level of obtained GRTS.

\noindent\textbf{Algorithm Description}

%\begin{minipage}%{12cm}
\begin{algorithm}%[H]
	\caption{The shape generalization algorithm.}\label{merge}
	\begin{algorithmic}[1]
		\Procedure{generalization}{$X$} %$S,l_i,EP,JP,\beta,threshold$
		\State Initialize $\mathcal{C}=\left\{ GRTS_1,\cdots,GRTS_N\right\}$ and $\theta=N$, where $C_i = \left\{ S_i\right\}_{i=1}^N$.
		\While{$\theta > 1$}
		\State Find the closed two common structures $GRTS_{\hat{j}}$ and $GRTS_{\hat{k}}$, $(\hat{i},\hat{k}=1,2,\cdots,\theta,\hat{i}\neq\hat{k})$.
		\State Merge $GRTS_{\hat{j}}$ and $GRTS_{\hat{k}}$ into one: $\hat{GRTS}=GRTS_{\hat{j}}\cup GRTS_{\hat{k}}$.
		\State Update $\mathcal{C}:\mathcal{C}\setminus GRTS_{\hat{j}}, \mathcal{C}\setminus GRTS_{\hat{k}},\mathcal{C}=\mathcal{C} \cup \hat{GRTS}, \theta = \theta - 1$.
		\EndWhile
		\State \Return{$\hat{GRTS}$}

		\EndProcedure
	\end{algorithmic}
\end{algorithm}
%\end{minipage}

\begin{figure*}[tb]
	\begin{center}
		\hspace{12pt}
		\includegraphics[width=0.98\linewidth]{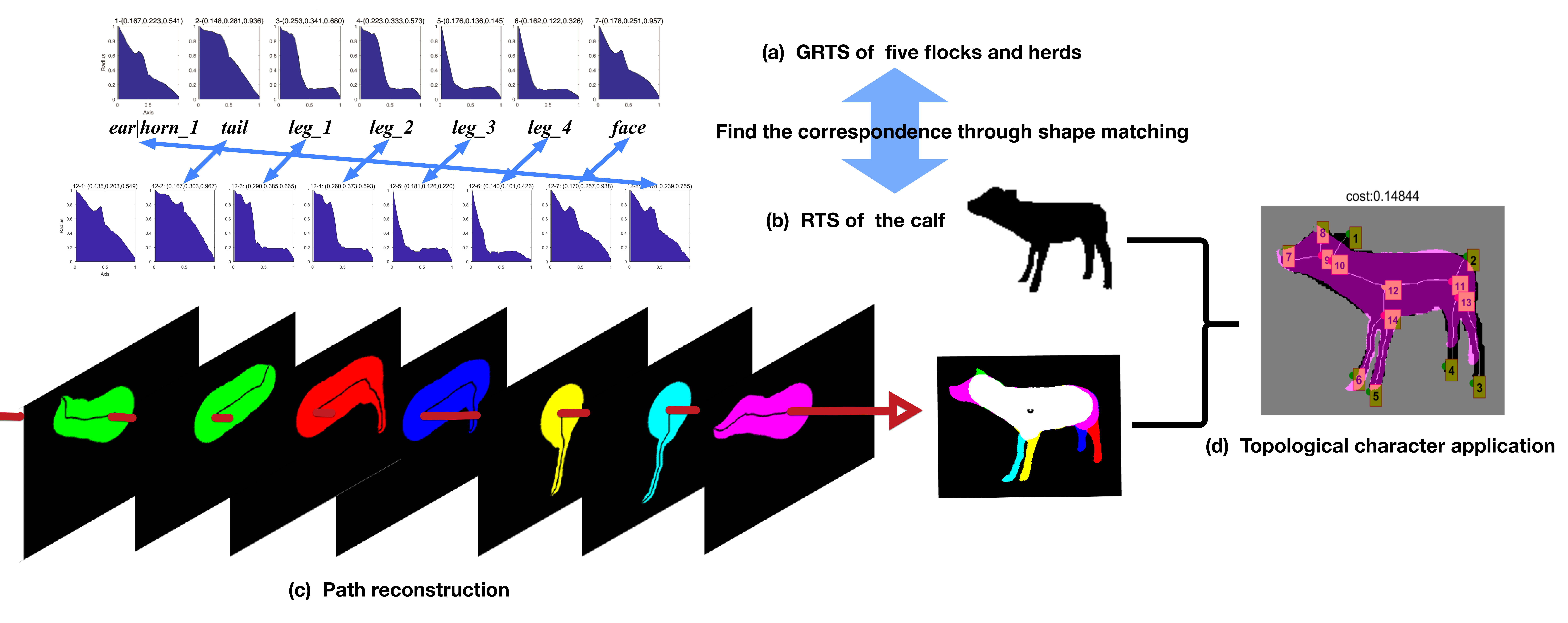}
	\end{center}
	%\vspace{-1.0em}
	\caption{The specific process of topological character application: (a) We first get the most general representation of the five cattle examples based on the generalization framework based on inductive learning, namely, the GRTS. (b) We find a specific example and get its RTS. (c) Correspondence between the GRTS and RTS is obtained according to our shape-matching algorithm mentioned in Section.~\ref{sec:simi}. (d) The path reconstruction is performed according to the feature vector of the GRTS through the correspondence, and the specific application of the GRTS in the instance is obtained.}
	
%	The specific process of topological character application: (a) We first get their most general representation on the five cattle examples based on the generalization framework based on inductive learning, namely GRTS. (b) We find a specific example, such as a cattle, gets its RTS. (c) Correspondence between GRTS and RTS is obtained according to our shape matching algorithm in Section.~\ref{sec:simi}.The path reconstruction is performed according to the feature vector of the GRTS through the correspondence, and the specific application of the GRTS in the instance is obtained, that is, (d).
%	}
%概念应用的具体过程：(a)我们首先根据基于归纳学习的泛化框架在五个牛羊实例上得到他们的最一般表示即GRTS. (b)我们找到一个具体的实例，如一头小牛，得到其RTS。(c)根据我们在Section.~\ref{sec:simi}的形状匹配算法得到GRTS和RTS的对应关系。通过对应关系根据GRTS的特征向量进行路径重建，得到GRTS在实例的具体应用，即(d)。}
	\label{fig:capp}
	\vspace{-.6em}
\end{figure*}

\subsection{Topological Character Application}
As shown in Figure.~\ref{fig:capp}(a), the GRTS contained a total of seven interconnected parts. The mass distribution of each local structure is shown in (a), and contains three important metric features of M, L, and vecdis (the straight square above the figure). These seven parts correspond to the limbs, head, tail, and horns (and possibly ears) of quadrupeds such as cattle and sheep. We used the skeleton path for reconstruction to clearly see the most summarized GRTS. In general, it was possible to carry out an abstract representation of the structure of seeking a common ground on quadrupeds such as cattle and sheep.

We could clearly see the difference between two objects based on their GRTS. Based on GRTS, we could not only get the label of the class, more importantly, we could know the difference on structure between shapes and even the small difference of the outline. This was not just a perceptual experience, but a rational understanding.

The specific process of topological character application was as follows. We got an object $X$ with $M$ end nodes and
%An object $X$ based on RTS representation has $M$ end nodes expressed as: …. 
the obtained GRTS with N end nodes. Based on the shape matching method in Section.~\ref{sec:simi}, we could obtaine the correspondence between two GRTSs. Then, using the coordinates of the end skeleton path of the corresponding end node of object $X$, the corresponding skeleton path of the GRTS was reconstructed to obtain an application of the topological character. We could clearly see the similarities and differences between the instance and our general representations of a generic class when applied to an object. The specific process is shown in Figure.~\ref{fig:capp}, with the final result in Figure.~\ref{fig:capp}(d). We can clearly see the similarities and differences between this specific example and the obtained GRTS. From the GRTS the most general characterization of the five cattle and sheep samples summed up the seven most characteristic parts of the body, namely, the limbs, face, horn (or ear), and tail. On a specific cattle sample, we found the correspondence between the GRTS and cattle RTS, and then reconstructed the skeleton path to obtain an image with a mask, as shown in the figure. The mask was the specific application of the GRTS in the instance. The specific differences between the GRTS and the cattle sample are clear. It can be seen that we obtained only one horn in the GRTS whereas the specific example had two horns. Moreover, from the ratio of the length of the two hind legs to the length of the body, the proportion of the example was longer than the sum of the GRTS (since we generalized the general characteristics of the samples, the proportion of the hind legs of the adult bull and the sheep to the body compared to the cattle To be small). This is a subtle difference that the human eye may not catch, but our algorithm could clearly see this difference.

\section{Experiment}
If the above method for shape topological character acquisition is effective, then it should be able to obtain typical topological and geometric features of the object from a small number of standard positive samples. GRTS could be used to analyze unseen samples. Thus give an explanation of whether it belongs to or not. In order to verify that the characterization we have rightly characterizes the common features of certain types of objects, we believe that the generalization of cross-datasets is necessary. In this section, we evaluated the proposed method using 3 datasets: Tari56 \cite{asian2005axis} and Kimia99 \cite{sebastian2004recognition}.

\subsection{Datasets}

\begin{figure}[!htb]

	\begin{center}
		\hspace{12pt}
		\includegraphics[width=0.65\linewidth]{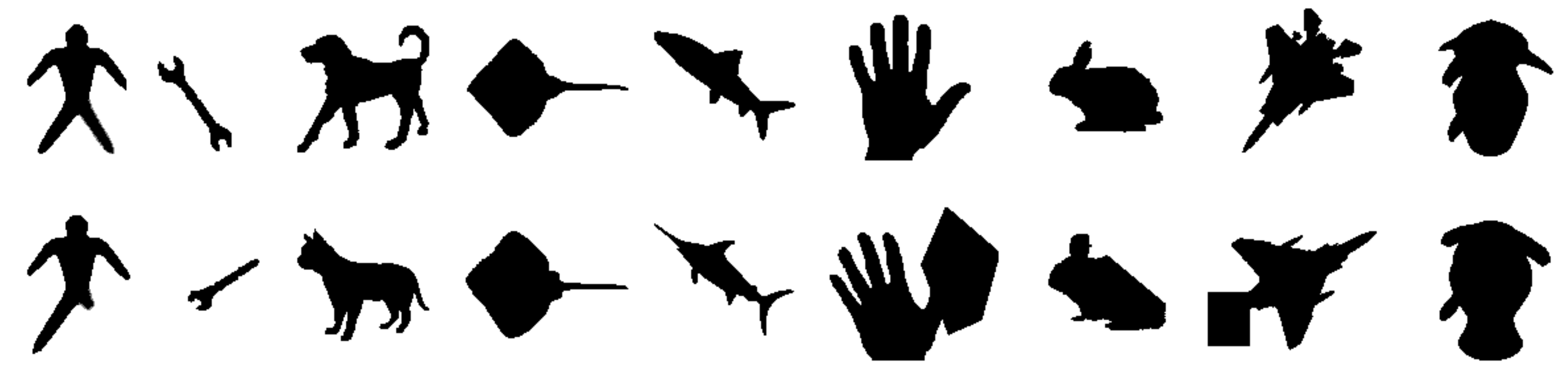}
	\end{center}
	%\vspace{-1.0em}
	\caption{Sample shapes from Kimia99 \cite{sebastian2004recognition} database.}
	\label{fig:kimia99imgs}
	\vspace{-.6em}
\end{figure}

Kimia99 Dataset: This database \cite{sebastian2004recognition} has images of 9 categories of objects, with 11 images per category for a total of 99 images (Figure.~\ref{fig:kimia99imgs}). It contains silhouettes of rabbits, quadrupeds, men, airplanes, fish, hands, rays, tools, and a miscellaneous class.

\begin{figure}[!htb]
	\begin{center}
		\hspace{12pt}
		\includegraphics[width=0.65\linewidth]{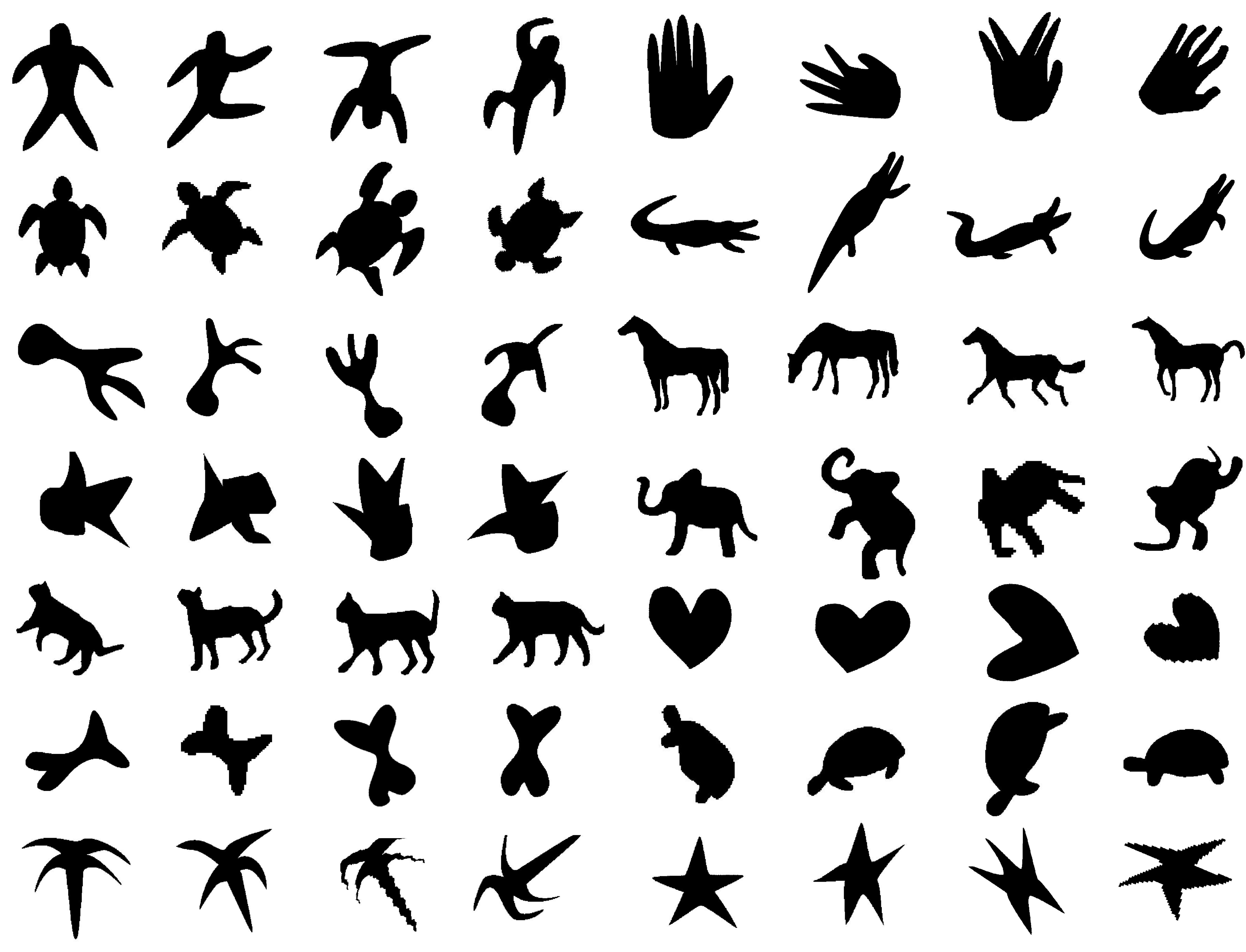}
	\end{center}
	%\vspace{-1.0em}
	\caption{All shapes from the Tari56 \cite{asian2005axis} database.}
	\label{fig:tari56imgs}
	\vspace{-.6em}
\end{figure}

Tari56 Database: The Tari56 database \cite{asian2005axis} is used for testing performance on non-rigid objects. It includes 14 classes of articulated shapes with 4 shapes in each class (Figure.~\ref{fig:tari56imgs}). No result on the whole database was presented in the original work. Moreover, there is no detailed explanation of retrieval results on this dataset using inner distance \cite{ling2007shape}, path similarity \cite{bai2008path}, or shape context \cite{Belongie2002Shape} methods. In this paper, we used the results in \cite{yang2016object}.

\subsection{Robustness of Recognition}
\label{sec:exprecog}
\begin{figure*}[t]
	\begin{center}
		\hspace{12pt}
		\includegraphics[width=0.80\linewidth]{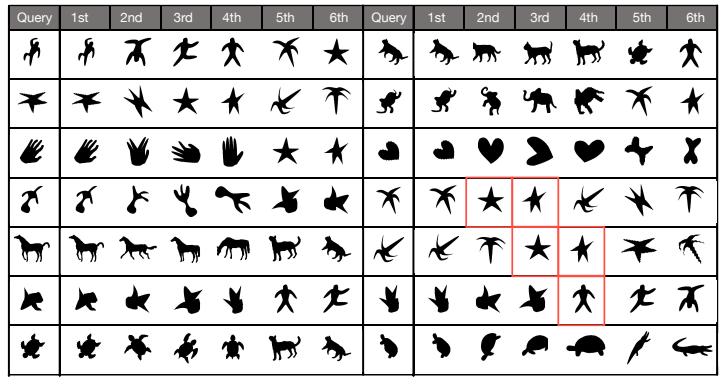}
	\end{center}
	%\vspace{-1.0em}
	\caption{Selected results of the proposed method on the Tari56 dataset. Since each class is composed of 4 shapes, the query and the first 4 most similar shapes should be in the same class. Red boxes mark the results where this is not the case.
	}
	\label{fig:res_tari56}
	\vspace{-.6em}
\end{figure*}

After getting a common feature about the shape of a certain type of object, the most basic verification was object recognition. To evaluate the recognition performance of our method, we tested it on the Tari56 dataset as shown in Figure. ~\ref{fig:tari56imgs}. We used each shape in this dataset as a query. Several representative results are shown in Figure.~\ref{fig:res_tari56}, where the six most similar shapes are shown for the queries. For each query, a perfect result should have the four most similar shapes in the same class as the query. The red squares indicate all the results where this is not the case. There were only 5 errors in a total of 224 query results, and the recognition rate on this dataset was 97.8\%. Using the bulls-eye test, the recognition rate was 99.4\%. Moreover, we can see that the wrong results were very similar to the query. For this dataset, we used parameters $\beta_1=30$ and $\beta_2=0.6$. Furthermore, because the Tari56 dataset contains primarily non-rigid objects, the value of $\beta_2$ is small.

\renewcommand\arraystretch{1.05}
\setlength{\tabcolsep}{5pt}
\begin{table}[!htb]
\begin{center}
\small
%\resizebox{1.0\linewidth}{!}{
\begin{tabular}{c|c|c|c|c|c}
%\begin{tabular}{p{1.5cm}<{\centering} | p{1.5cm}<{\centering} | p{1.5cm}<{\centering} | p{1.5cm}<{\centering} | p{1.5cm}<{\centering} | p{1.5cm}<{\centering}}
\hline
Tari56 & 1st & 2nd & 3rd & 4th & Accuracy($\%$)\\
\hline
ID \cite{ling2007shape} & 56 & 46 & 37 & 28 & 74.6\\
SC \cite{Belongie2002Shape} & 52 &17 & 10 & 10 & 39.7\\
PS \cite{bai2008path} & 56 & 49 & 44 & 40 & 84.4\\
HS \cite{yang2016object} & 56 & 51 & 50 & 33 & 84.8\\
ours(without SPA) & 56 & 55 & 53 & 53 & 96.9\\
\textbf{ours} & \textbf{56} & \textbf{55} & \textbf{54} & \textbf{54} & 97.8\\
\hline
\end{tabular}
%}
\end{center}
\caption{Table 1. Results comparison on Tari56 dataset.}%, and our 5 results are \{6.64, 6.43, 6.61, 6.53, 6.86\}.
\label{tab:tari56_res}
\end{table}

As shown in Table.~\ref{tab:tari56_res}, we achieved the best results among all other methods. When we did not use the SPA but set parameter $\beta_2=0$, we still obtained a 96.9\% correct rate. This showed that our method can well represent the structural information of the object.

\begin{figure*}[!htb]
	\begin{center}
		\hspace{12pt}
		\includegraphics[width=0.92\linewidth]{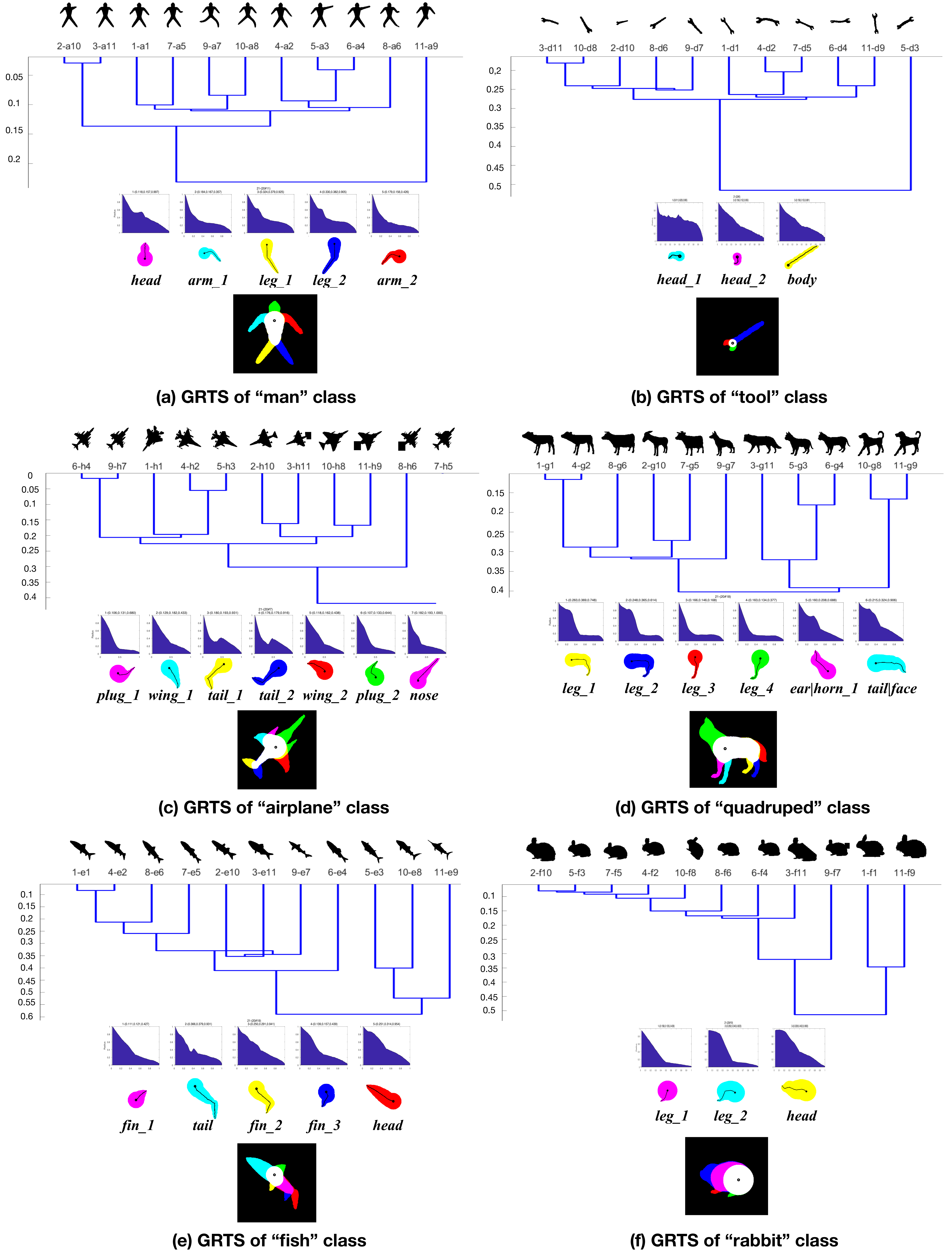}
	\end{center}
	%\vspace{-1.0em}
	\caption{GRTS experiments on Kimia-99 database.}
	\label{fig:ck6}
	\vspace{-.6em}
\end{figure*}

\begin{figure*}[!htb]
	\begin{center}
		\hspace{12pt}
		\includegraphics[width=0.92\linewidth]{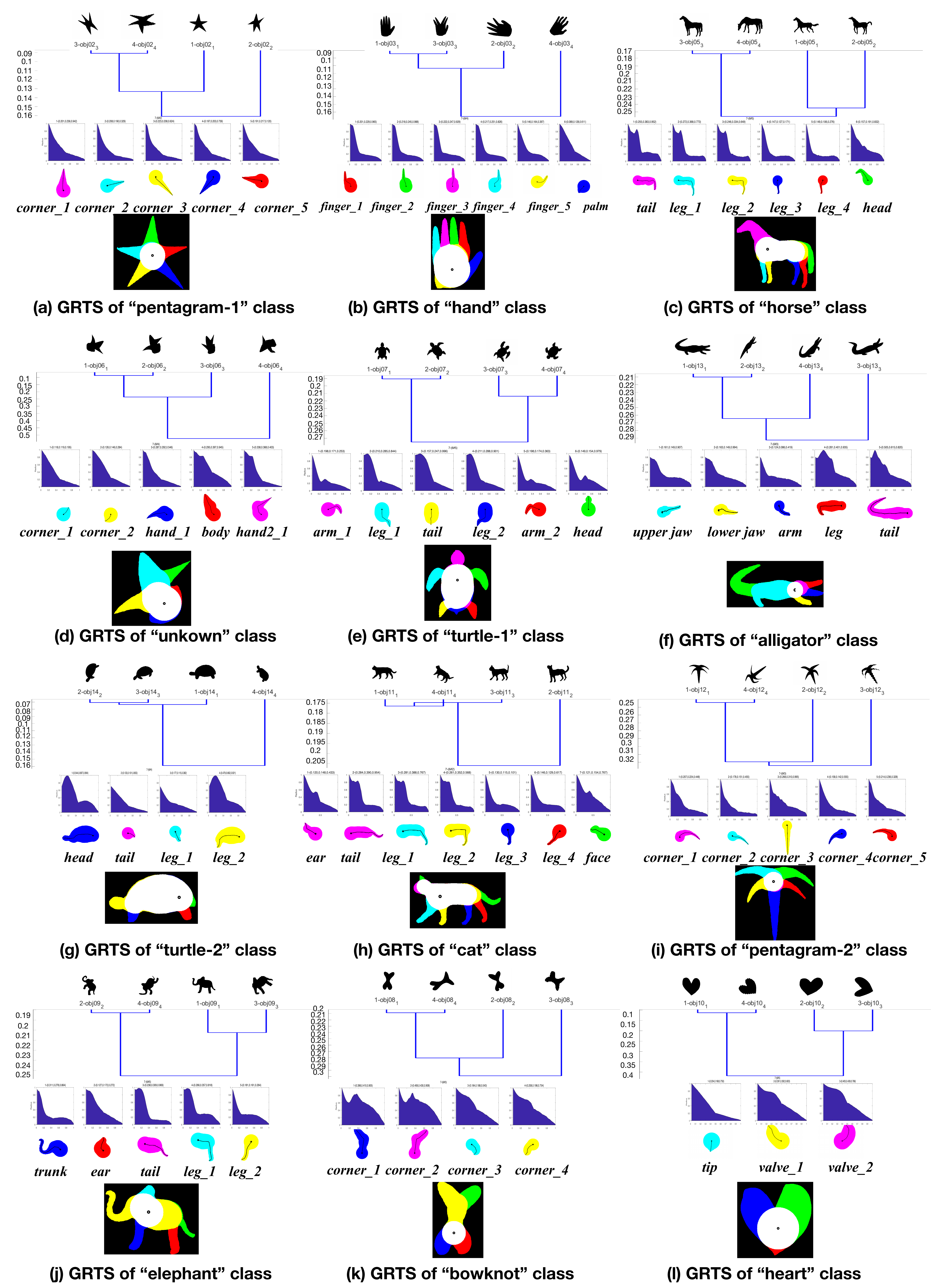}
	\end{center}
	%\vspace{-1.0em}
	\caption{GRTS experiments on Tari56 database.
%GRTS Experiments on Tari56 database.
	}
	\label{fig:ct12}
	\vspace{-.6em}
\end{figure*}

\subsection{GRTS can Characterize the Essential Configuration Features of Objects}
We conducted inductive learning on some of the data of Kimia99 and Tari56 datasets, and showed the obtained GRTS. And from the form of direct observation of the geometric meaning of the most general representation we get on the same kind of objects. The parameters on Kimia-99 were set to $\beta_1 = 29$ and $\beta_2=0.7$. Since the Kimia-99 dataset is relatively fixed compared to the Tari56 dataset, the value of $\beta_2$ is slightly larger. And since kimia-99 is more involved in occlusion, missing, etc. of objects, the value of $\beta_1$ is smaller.

We performed experiments with the same GRTS on the Kimia-99 and Tari-56 datasets, as shown in Figure.~\ref{fig:ck6} and Figure.~\ref{fig:ct12}. We can clearly see the generalization process of objects of the same class and the resulting GRTSs. In order to visually see the meaning of the histogram in the GRTS, we reconstructed the skeleton path of the same kind of object. More importantly, the GRTS could characterize the most general structural features of objects of the same kind, as in the GRTS experiment of 'quadruped' category (labeled g) on the Kimia99 dataset (Figure.~\ref{fig:ck6}(d)). For quadrupeds with large internal distances, such as cats, foxes, and dogs, our GRTS could summarize their most general essential features, such as four limbs, at least one horn (or ear), and tails.

There is a very interesting phenomenon here. We observe the generalized hierarchical tree of quadrupeds, and can clearly see the generic relationship of objects: cattle and sheep occupy a large branch, while cats and foxes occupy a small branch. All quadrupeds were eventually generalized into one category. Our generalized hierarchical tree can reflect the true generic relationship of objects to some extent.

As shown in Figure.~\ref{fig:ck6}(c), the 'airplanes' category contains various forms: the number of plug-ins varies, and the shape of the tail is different. However, our final GRTS result ignored the individualized structure of each instance. It also integrated small differences between components, such as larger tails and smaller ones. Our GRTS could characterize the most common structure of the airplanes: two wings, two tails, a nose, and two external components. All of the above experiments showed that our GRTS could gradually summarize the most essential configuration features learned through an RTS-based generalization framework.

\begin{figure}[!htb]
	\begin{center}
		\hspace{12pt}
		\includegraphics[width=0.82\linewidth]{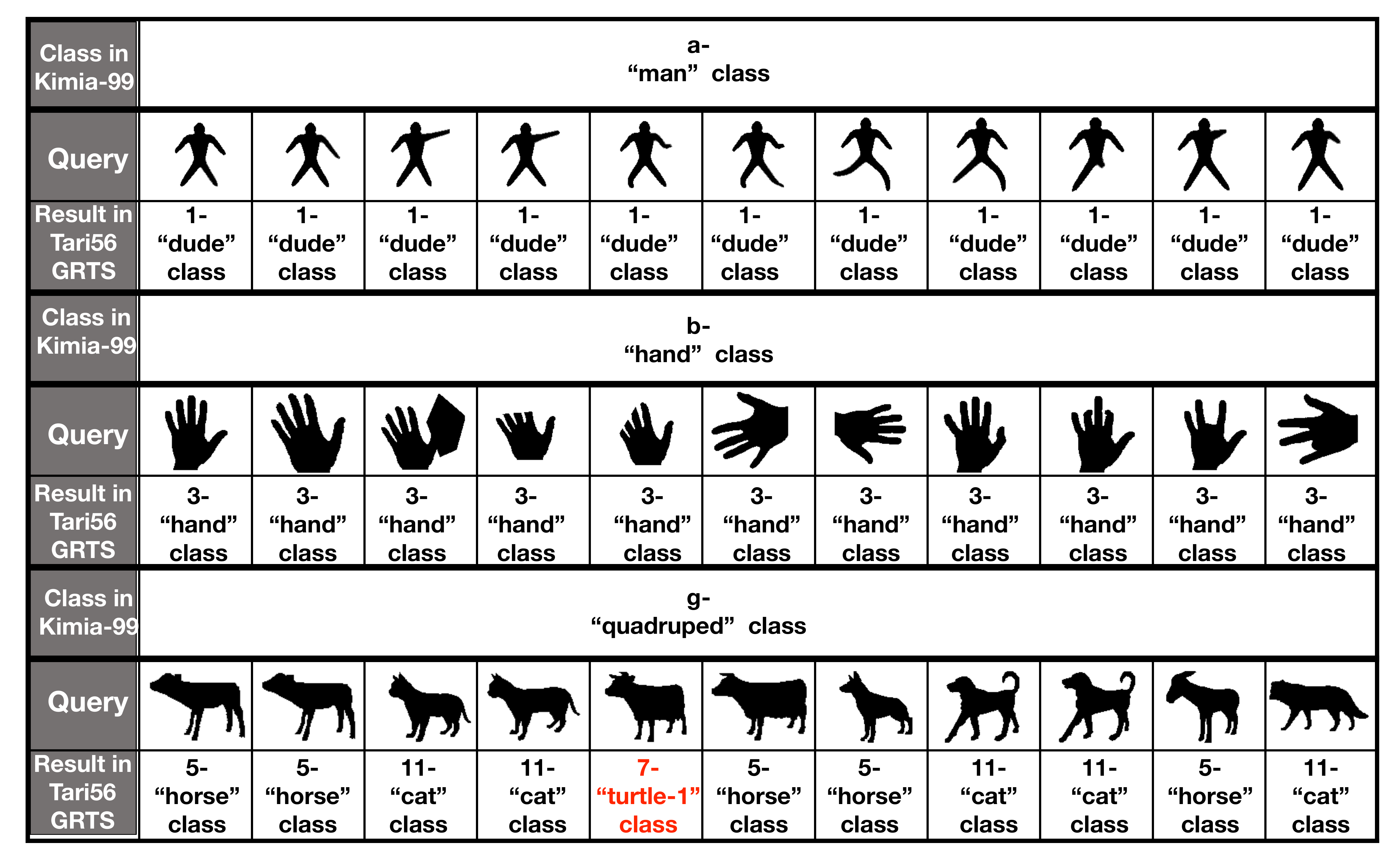}
	\end{center}
	%\vspace{-1.0em}
	\caption{GRTS generalization performance verification experiment 1. We used the Kimia-99 dataset object to retrieve and return the GRTS category with the highest similarity in the Tari56 dataset. Red indicates a wrong result.
%	GRTS Generalization Performance Verification Experiment 1. We use the Kimia-99 dataset object to retrieve and return the GRTS category with the highest similarity in the Tari56 dataset. Red indicates the wrong result. 
	}
%	GRTS泛化性能验证实验1. 我们用Kimia-99数据集中与Tari56数据集中物体种类有交集的几类物体作为检索，返回Tari56中与之相似度最大的GRTS。红色表示错误结果。}
	\label{fig:grts_k}
	\vspace{-.6em}
\end{figure}

\subsection{Generalization Performance Verification Across Datasets}
\begin{figure}[!htb]
	\begin{center}
		\hspace{12pt}
		\includegraphics[width=0.88\linewidth]{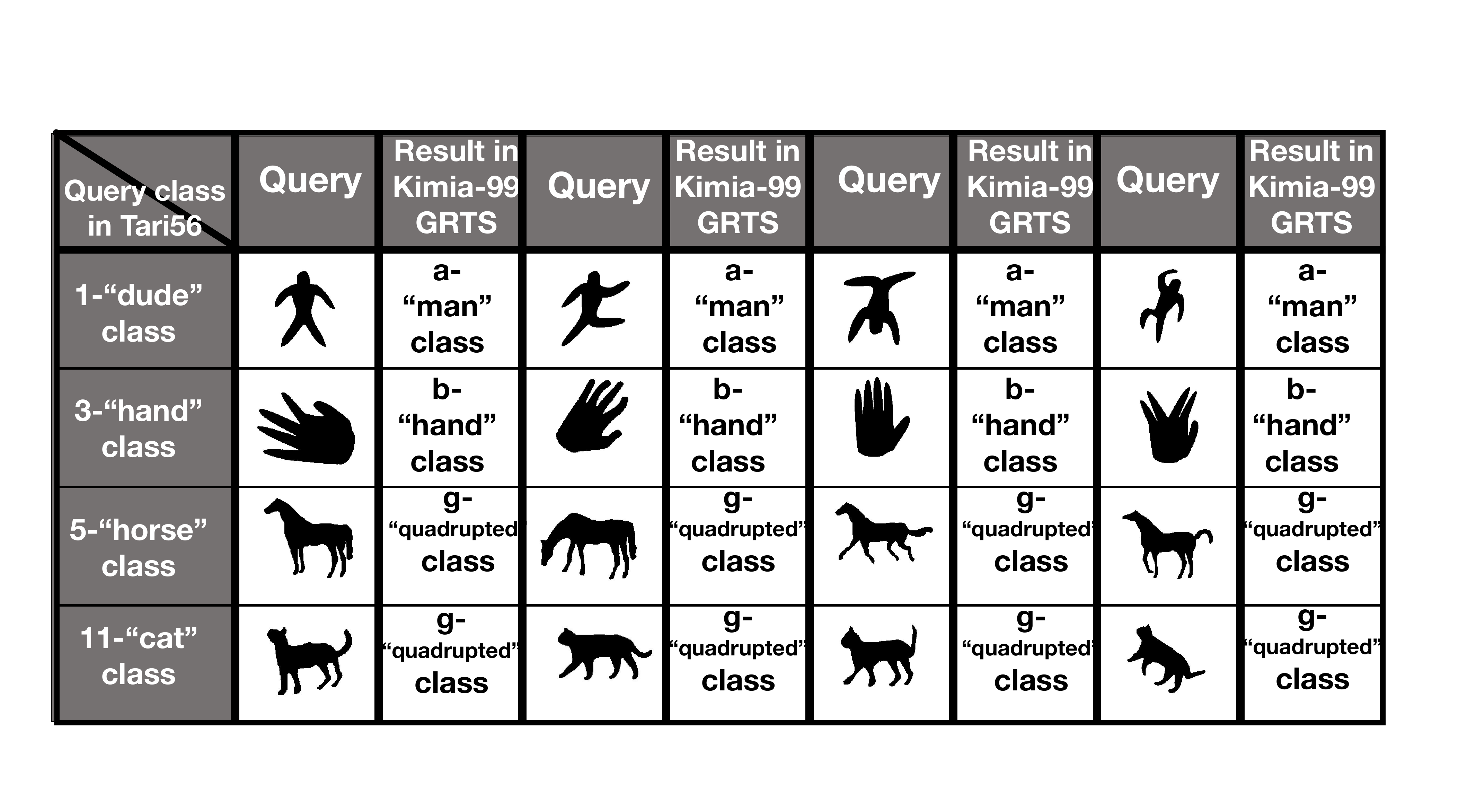}
	\end{center}
	%\vspace{-1.0em}
	\caption{GRTS generalization performance verification experiment 2. We used the objects in the Tari56 dataset to retrieve and return the categories of GRTS with the highest similarity in the Kimia-99 dataset.
%	GRTS generalization performance verification experiment 2. We use the objects in the Tari56 dataset to retrieve and return the categories of GRTS with the highest similarity in the Kimia-99 dataset. 
	}
%	GRTS泛化性能验证实验2. 我们用Tari56数据集中的部分物体作为检索，返回Kimia-99数据集中与之相似度最大的GRTS。}
	\label{fig:grts_t}
	\vspace{-.6em}
\end{figure}

To demonstrate the generalization performance of the resulting GRTSs, we performed cross-classification recognition experiments on the Kimia-99 and Tari-56 datasets, respectively. We used the samples from dataset A to retrieve the characterizations learned from dataset B to find the most sympathetic representation. The two datasets are completely disjoint, and the test data was completely new to the GRTS. If the GRTS on dataset B could be correctly classified for the samples in dataset A, our method’s generalization performance was good. The results were shown in Figure.~\ref{fig:grts_k} and Figure.~\ref{fig:grts_t}.

\begin{figure}[tb]
	\begin{center}
%		\hspace{12pt}
		\includegraphics[width=0.90\linewidth]{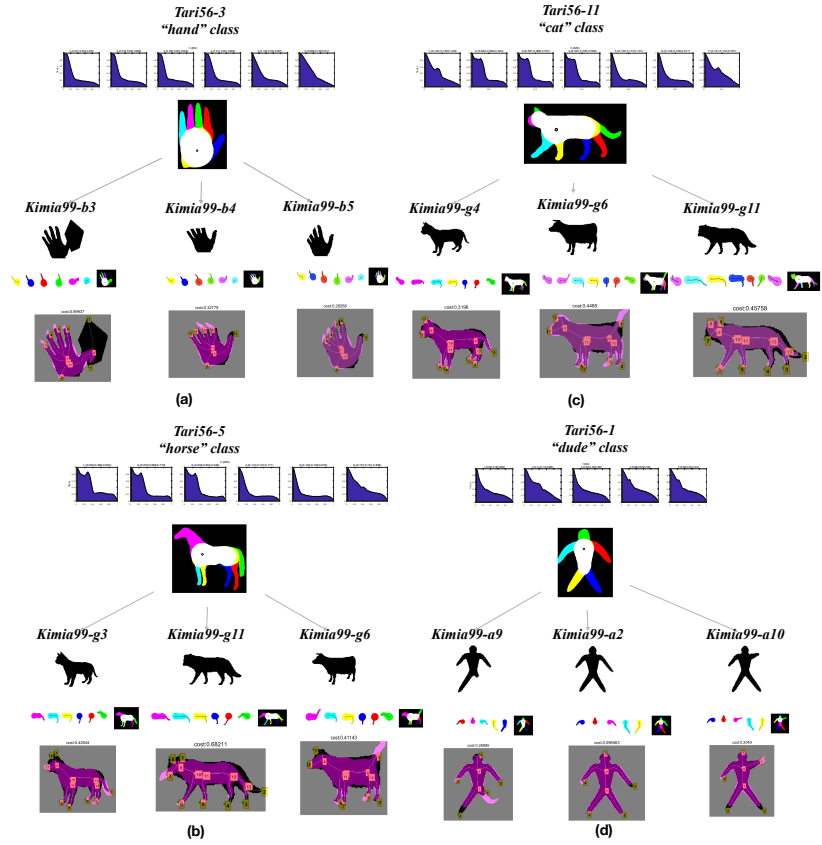}
	\end{center}
	%\vspace{-1.0em}
	\caption{Topological character application experiments on the Kimia-99 database using Tari56 GRTS.
%	Topological character application experiments on Kimia-99 database using Tari56 GRTS.
	}
	\label{fig:jieshi}
	\vspace{-.6em}
\end{figure}

\subsection{Interpretability Verification}
The reason why we think that simplifying the identification problem into a classification problem is flawed is that, although it gives a category label, the classification algorithm does not give a reason why a sample belongs to or does not belong to the category. Classification is a crude approach that does not explain why there is a lack of explicit representation to support a detailed analysis of the image. The GRTS in this article can help the computer explain the recognition conclusion, which can be verified and validated. In the following, we will summarize the GRTS learned from the topological character application on the new test sample, verify the interpretability of our characterization, and see the difference between the test sample and the general characterization so that we can find the part of the test sample that is redundant or missing. This can form two interpretations: 'being occluded' or 'being truncated.'

As shown in Figure.~\ref{fig:jieshi}, we used the GRTS of several classes obtained from the Tari56 dataset to apply the topological character to a sample of Kimia99's closest category to discover the specific differences of the new test sample relative to the GRTS.

Figure.~\ref{fig:jieshi}(a) shows the result of the topological character application of several examples in Tari56's 'hand' category in Kimia99. Kimia99-b3 is a hand with huge occlusion, and after applying the 'hand' category GRTS, we could clearly see the difference between the two hands: the original image is more redundant than the purple mask area. (purple mask was the image we generated using the GRTS, and the black base map is the original picture). The middle and right images show the results of the application on Kimia99-b4 and b5; the purple mask generated a complete finger above the missing part of the original image. In addition, the top part of the figure presents the similarity recognized by the GRTS. The left picture shows that the matching cost was obviously improved despite the existence of huge obstructions. The matching cost of the middle and right images was very low, indicating that the more they matched the general representation pattern of the hand, the more they look like the hand. This shows that our GRTS not only recognized the finger but also had a strong interpretability that tells us occluded and missing details.

Figure.~\ref{fig:jieshi}(b) is the result of a partial sample of the GRTS application of the 'horse' category of Tari56 to the Kimia99 'quadruped' category. The left image shows the difference between the cat’s and horse's general pattern. For example, the horse's tail is thicker and longer than the cat's, and the horse's face is longer than the cat's. The middle image shows that the fox's tail is larger and heavier than the horse's and the horse's face is much longer than the fox’s. In the right image it can be seen that the cow has more horns than the horse, and the tail of the cattle is short and thick while the horse’s tail is more prominent. Figure.~\ref{fig:jieshi}(c) shows that the cat's face is rounder and shorter than that of the cattle, and its mouth is not as pointed as the fox's. Furthermore, the cat's tail is smaller and more prominent than that of the cattle. From the left of Figure.~\ref{fig:jieshi}(d) we could see that the instance belongs to the 'man' category; however, the left leg is obviously missing, and the pattern generated by the GRTS can explore this difference. Another example is the subtle difference in the length of the limbs of the villain in the various pictures.

All of the above experimental results showed that our RTS characterization method is highly interpretable, not only to explore the similarity between objects, but also to discover differences.

\begin{figure*}[tb]
	\begin{center}
%		\hspace{12pt}
		\includegraphics[width=0.95\linewidth]{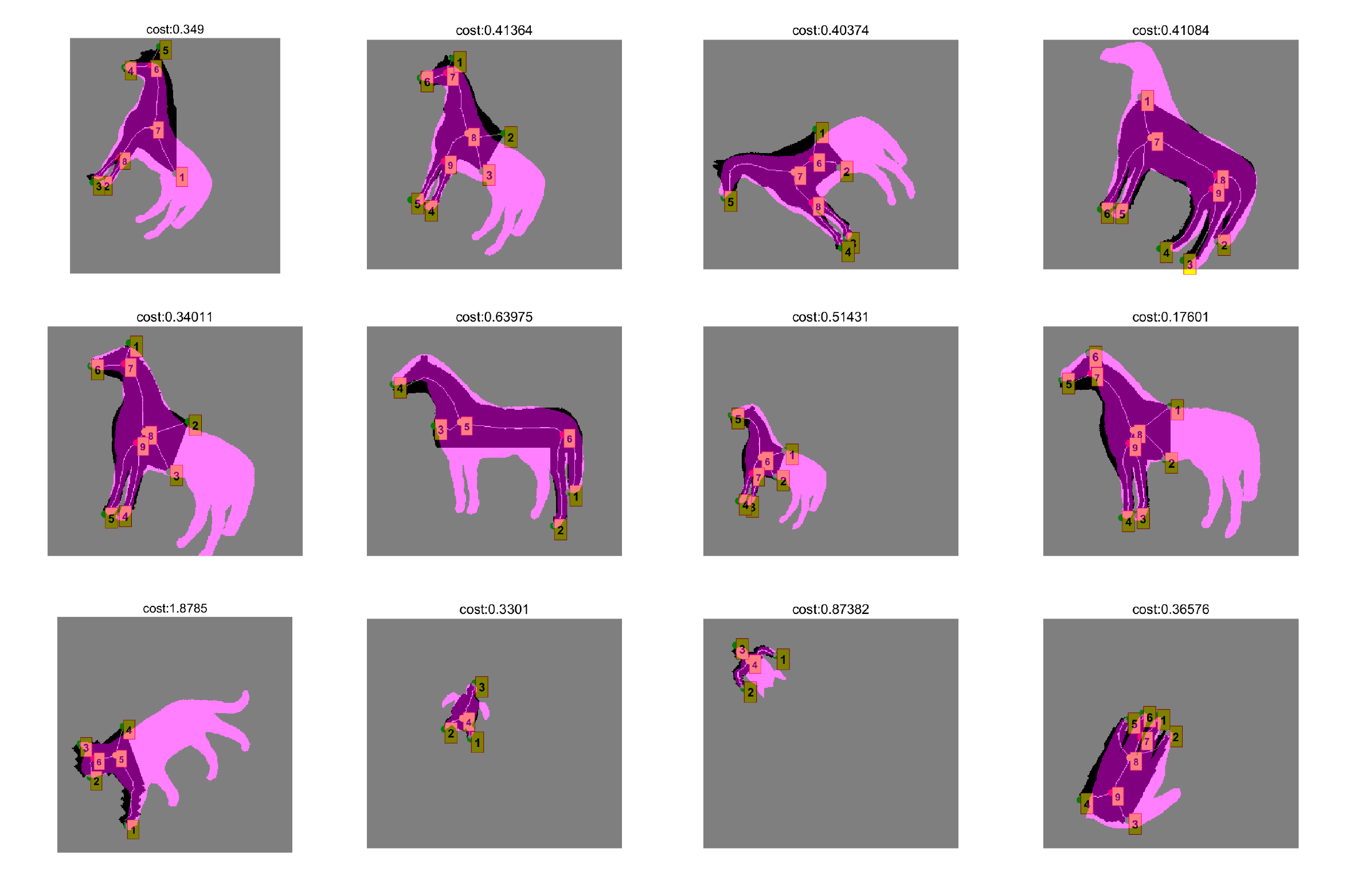}
	\end{center}
	\vspace{-1.0em}
	\caption{Active processing results.}
	\label{fig:ap}
	\vspace{-.6em}
\end{figure*}

\subsection{Active Processing}
Cognitive psychology divides the human pattern recognition process into two iterative phases: the data-driven phase and the concept-driven phase. These are also known as bottom-up processing and top-down processing. The research on the latter stage is less theoretically or engineeringly, because it involves many problems, such as how the high-level concept is characterized, how the upper layer exerts influence on the lower layer, and how the underlying features match the high-level semantic representation. The so-called active processing is aimed at the top-down stage. Specifically, it refers to how to rediscover, re-screen, and reorganize the physical stimulation of the lower layer according to the semantic concept of the upper layer. Since this is a process with a clear pre-set target that actively collects evidence to validate a hypothesis, it does not passively accept the pre-processing results; however, it can actively correct bad pre-processing. This active processing is an advanced stage of brain recognizing object Since a brief description of the shape of the object was obtained in the early stage of the present study, it was possible to use this formal characterization to guide the analysis of an original image. For example, one form of active processing is to 'complement the image,' which is to identify an object based on the visible local information and then complete the missing portion to form a complete shape about the object. In fact, the purpose of doing this is not only to see a complete shape, but basing on a more complete image to verify and validate the correctness of partial image identification. In general, classifier-based object recognition methods are not capable of such active processing because the classifier does not contain explicit representations of certain types of objects.

We will show the experimental results of active machining through several examples. Figure.~\ref{fig:ap} shows a horse with a large portion of its shape missing. However, according to the human pattern recognition process, we can still identify a part of the horse, and thus can create the complete horse shape in the brain based on the information that is available. We first obtained the GRTS representation of the horse. Then, based on the shape-matching algorithm presented in Section 4, we found the endpoint matching relationship between the missing horse shape and the GRTS, and applied the similar transformation (only two pairs of matching points were needed). The similar transformation matrix was obtained. The rest of the GRTS was processed onto the missing horse shape to complete the active machining process.

\section{Conclusion}
In this paper, we proposed a formal representation of a topological structure based on an object's skeleton (RTS), and introduced a 'spine-like axis' (SPA) to further constrain the spatial distribution of the object’s structure. Experimental results showed that this method can effectively characterize the structural features of objects. Based on the RTS, we replaced the Euclidean distance with the Fr{\'e}chet distance to improve the previous skeleton path similarity measurement method. The OSB algorithm was used to elastically match the object structure to obtain the structural matching relationship between objects.

This paper proposed inductive learning on RTS characterization for topological character acquisition. The GRTS could derive declarative knowledge to describe the essential characteristics of objects. Experiments proved that this formal representation itself contains moderate details, has good usability, can effectively extract the explicit meaning representation of categories, and facilitates the active processing driven by knowledge.

Inductive learning is one of the most important ways for humans to acquire new knowledge. Only a small number of representative samples are required to discover their common features and form the most general description of objects. The GRTS in this paper has such capabilities. Explicitly expressing the laws discovered by learning methods in a declarative and general way is inevitable when upgrade to some level of rationalism.

Cognitive psychology shows that the principle of holistic priority is one of the most reliable and effective means for us to cope with the task of understanding complex scenes. In this paper, inductive learning was based on the topological structure of an object’s shape, and was not limited to the underlying and local physical features but encompassed a holistic description. The experimental results of this paper reflect the principle of global precedence.

\section*{Acknowledgements}
This work was supported by the NSFC Project (Project Nos. 61771146 and 61375122), (in part) by Shanghai Science and Technology Development Funds (Project Nos. 13dz2260200, 13511504300). We thank LetPub (www.letpub.com) for its linguistic assistance during the preparation of this manuscript.

\section*{References}

\bibliography{mybibfile}

\begin{thebibliography}{10}
\expandafter\ifx\csname url\endcsname\relax
  \def\url#1{\texttt{#1}}\fi
\expandafter\ifx\csname urlprefix\endcsname\relax\def\urlprefix{URL }\fi
\expandafter\ifx\csname href\endcsname\relax
  \def\href#1#2{#2} \def\path#1{#1}\fi

\bibitem{Dueker2005Infants}
G.~Dueker, A.~Needham, Infants' object category formation and use: Real-world
  context effects on category use in object processing, Visual Cognition 12~(6)
  (2005) 1177--1198.

\bibitem{Bloom2004Descartes}
P.~Bloom, Descartes' baby: How the science of child development explains what
  makes us human., Basic Books 11~(9) (2004) 89--91.

\bibitem{gopnik1999scientist}
A.~Gopnik, A.~N. Meltzoff, P.~K. Kuhl, The scientist in the crib: Minds,
  brains, and how children learn., William Morrow \& Co, 1999.

\bibitem{Shinskey2014Picturing}
J.~L. Shinskey, L.~J. Jachens, Picturing objects in infancy., Child Dev 85~(5)
  (2014) 1813--1820.

\bibitem{brincat2004underlying}
S.~L. Brincat, C.~E. Connor, Underlying principles of visual shape selectivity
  in posterior inferotemporal cortex, Nature neuroscience 7~(8) (2004) 880.

\bibitem{Ito1995Size}
M.~Ito, H.~Tamura, I.~Fujita, K.~Tanaka, Size and position invariance of
  neuronal responses in monkey inferotemporal cortex., Journal of
  Neurophysiology 73~(1) (1995) 218.

\bibitem{Majaj2015Simple}
N.~J. Majaj, H.~Hong, E.~A. Solomon, J.~J. Dicarlo, Simple learned weighted
  sums of inferior temporal neuronal firing rates accurately predict human core
  object recognition performance, Journal of Neuroscience 35~(39) (2015)
  13402--13418.

\bibitem{Bruce1981Visual}
C.~Bruce, R.~Desimone, C.~G. Gross, Visual properties of neurons in a
  polysensory area in superior temporal sulcus of the macaque., Journal of
  Neurophysiology 46~(2) (1981) 369--84.

\bibitem{tsao2003faces}
D.~Y. Tsao, W.~A. Freiwald, T.~A. Knutsen, J.~B. Mandeville, R.~B. Tootell,
  Faces and objects in macaque cerebral cortex, Nature neuroscience 6~(9)
  (2003) 989.

\bibitem{Tsao2006A}
D.~Y. Tsao, W.~A. Freiwald, R.~B. Tootell, M.~S. Livingstone, A cortical region
  consisting entirely of face-selective cells., Science 311~(5761) (2006)
  670--674.

\bibitem{Lin2007Neural}
L.~Lin, G.~Chen, H.~Kuang, D.~Wang, J.~Z. Tsien, Neural encoding of the concept
  of nest in the mouse brain, Proceedings of the National Academy of Sciences
  of the United States of America 104~(14) (2007) 6066--6071.

\bibitem{Fang2018Semantic}
Y.~Fang, X.~Wang, S.~Zhong, L.~Song, Z.~Han, G.~Gong, Y.~Bi, Semantic
  representation in the white matter pathway, Plos Biology 16~(4) (2018)
  e2003993.

\bibitem{Fernandino2015Concept}
L.~Fernandino, J.~R. Binder, R.~H. Desai, S.~L. Pendl, C.~J. Humphries, W.~L.
  Gross, L.~L. Conant, M.~S. Seidenberg, Concept representation reflects
  multimodal abstraction: A framework for embodied semantics, Cerebral Cortex
  26~(5) (2015) 2018.

\bibitem{martin2016grapes}
A.~Martin, Grapes—grounding representations in action, perception, and
  emotion systems: How object properties and categories are represented in the
  human brain, Psychonomic bulletin \& review 23~(4) (2016) 979--990.

\bibitem{frechet1906quelques}
M.~M. Fr{\'e}chet, Sur quelques points du calcul fonctionnel, Rendiconti del
  Circolo Matematico di Palermo (1884-1940) 22~(1) (1906) 1--72.

\bibitem{Belongie2002Shape}
S.~Belongie, J.~Malik, J.~Puzicha, Shape matching and object recognition using
  shape contexts, IEEE Transactions on Pattern Analysis \& Machine Intelligence
  24~(4) (2002) 509--522.

\bibitem{Latecki2000Shape}
L.~J. Latecki, R.~Lakamper, T.~Eckhardt, Shape descriptors for non-rigid shapes
  with a single closed contour, in: IEEE Conference on Computer Vision \&
  Pattern Recognition, 2000.

\bibitem{Sun2005Classification}
K.~B. Sun, B.~J. Super, Classification of contour shapes using class segment
  sets, in: IEEE Computer Society Conference on Computer Vision \& Pattern
  Recognition, 2005.

\bibitem{sivic2003video}
J.~Sivic, A.~Zisserman, Video google: A text retrieval approach to object
  matching in videos, in: null, IEEE, 2003, p. 1470.

\bibitem{csurka2004visual}
G.~Csurka, C.~Dance, L.~Fan, J.~Willamowski, C.~Bray, Visual categorization
  with bags of keypoints, in: Workshop on statistical learning in computer
  vision, ECCV, Vol.~1, Prague, 2004, pp. 1--2.

\bibitem{felzenszwalb2000efficient}
P.~F. Felzenszwalb, D.~P. Huttenlocher, Efficient matching of pictorial
  structures, in: Computer Vision and Pattern Recognition, 2000. Proceedings.
  IEEE Conference on, Vol.~2, IEEE, 2000, pp. 66--73.

\bibitem{felzenszwalb2003pictorial}
P.~F. Felzenszwalb, D.~P. Huttenlocher, Pictorial structures for object
  recognition, in: IJCV, Citeseer, 2003.

\bibitem{zahn1972fourier}
C.~T. Zahn, R.~Z. Roskies, Fourier descriptors for plane closed curves, IEEE
  Transactions on computers 100~(3) (1972) 269--281.

\bibitem{chuang1996wavelet}
G.-H. Chuang, C.-C. Kuo, Wavelet descriptor of planar curves: Theory and
  applications, IEEE Transactions on Image Processing 5~(1) (1996) 56--70.

\bibitem{davies2004machine}
E.~R. Davies, Machine vision: theory, algorithms, practicalities, Elsevier,
  2004.

\bibitem{batchelor1980hierarchical}
B.~Batchelor, Hierarchical shape description based upon convex hulls of
  concavities, Cybernetics and System 10~(1-3) (1980) 205--210.

\bibitem{blum1973biological}
H.~Blum, Biological shape and visual science (part i), Journal of theoretical
  Biology 38~(2) (1973) 205--287.

\bibitem{calabi1968shape}
L.~Calabi, W.~E. Hartnett, Shape recognition, prairie fires, convex
  deficiencies and skeletons, The American Mathematical Monthly 75~(4) (1968)
  335--342.

\bibitem{Fatih2009Skeletal}
D.~M. Fatih, A.~Shokoufandeh, S.~J. Dickinson, Skeletal shape abstraction from
  examples, IEEE Transactions on Pattern Analysis \& Machine Intelligence
  31~(5) (2009) 944.

\bibitem{Torsello2006Learning}
A.~Torsello, E.~R. Hancock, Learning shape-classes using a mixture of
  tree-unions, IEEE Transactions on Pattern Analysis \& Machine Intelligence
  28~(6) (2006) 954--67.

\bibitem{Wei2013Shape}
S.~Wei, W.~Yan, B.~Xiang, H.~Wang, L.~J. Latecki, Shape clustering: Common
  structure discovery, Pattern Recognition 46~(2) (2013) 539--550.

\bibitem{borgefors1986distance}
G.~Borgefors, Distance transformations in digital images, Computer vision,
  graphics, and image processing 34~(3) (1986) 344--371.

\bibitem{Wei2013Skeleton}
S.~Wei, B.~Xiang, X.~W. Yang, L.~J. Latecki, Skeleton pruning as trade-off
  between skeleton simplicity and reconstruction error, Science China
  Information Sciences 56~(4) (2013) 1--14.

\bibitem{eiter1994computing}
T.~Eiter, H.~Mannila, Computing discrete fr{\'e}chet distance, Tech. rep.,
  Citeseer (1994).

\bibitem{latecki2007optimal}
L.~J. Latecki, Q.~Wang, S.~Koknar-Tezel, V.~Megalooikonomou, Optimal
  subsequence bijection, in: icdm, IEEE, 2007, pp. 565--570.

\bibitem{bai2008path}
X.~Bai, L.~J. Latecki, Path similarity skeleton graph matching, IEEE
  transactions on pattern analysis and machine intelligence 30~(7) (2008)
  1282--1292.

\bibitem{asian2005axis}
C.~Asian, S.~Tari, An axis-based representation for recognition, in: Tenth IEEE
  International Conference on Computer Vision (ICCV'05) Volume 1, Vol.~2, IEEE,
  2005, pp. 1339--1346.

\bibitem{sebastian2004recognition}
T.~B. Sebastian, P.~N. Klein, B.~B. Kimia, Recognition of shapes by editing
  their shock graphs, IEEE Transactions on pattern analysis and machine
  intelligence 26~(5) (2004) 550--571.

\bibitem{ling2007shape}
H.~Ling, D.~W. Jacobs, Shape classification using the inner-distance, IEEE
  transactions on pattern analysis and machine intelligence 29~(2) (2007)
  286--299.

\bibitem{yang2016object}
C.~Yang, O.~Tiebe, K.~Shirahama, M.~Grzegorzek, Object matching with
  hierarchical skeletons, Pattern Recognition 55 (2016) 183--197.

\end{thebibliography}

\end{document}